%% file: main.tex
\documentclass{article}
\usepackage{ijcai21}

\usepackage{times}
\usepackage{soul}
\usepackage{url}
\usepackage[hidelinks]{hyperref}
\usepackage[utf8]{inputenc}
\usepackage[small]{caption}
\usepackage{graphicx}
\usepackage{amsmath}
\usepackage{amsthm}
\usepackage{booktabs}
\usepackage{algorithm}
\usepackage{algorithmic}

\usepackage{subfigure}
\input{math_commands.tex}

\newtheorem{theorem}{Theorem}
\newtheorem{lemma}[theorem]{Lemma}

\title{Robust Federated Learning with Attack-Adaptive Aggregation}

\author{
Ching Pui Wan\and
Qifeng Chen\\
\affiliations
The Hong Kong University of Science and Technology\\
\emails
\{cpwan, cqf\}@ust.hk
}

\begin{document}

\maketitle

\begin{abstract}
Federated learning is vulnerable to various attacks, such as model poisoning and backdoor attacks, even if some existing defense strategies are used. To address this challenge, we propose an attack-adaptive aggregation strategy to defend against various attacks for robust federated learning. The proposed approach is based on training a neural network with an attention mechanism that learns the vulnerability of federated learning models from a set of plausible attacks. To the best of our knowledge, our aggregation strategy is the first one that can be adapted to defend against various attacks in a data-driven fashion. Our approach has achieved competitive performance in defending model poisoning and backdoor attacks in federated learning tasks on image and text datasets. 
\end{abstract}

\section{Introduction}
\label{intro}
Federated learning allows multiple clients to collectively train a neural network without directly sharing their own private data \cite{McMahanMRHA17,smith2017federated}. The federated learning framework has been proposed for diverse applications such as mobile applications, healthcare, and financial assessment \cite{yang2019federated}. Despite the large potential of federated learning in real-life applications, it is vulnerable to numerous attacks, including data poisoning and model poisoning \cite{blanchard2017machine,BagdasaryanVHES20,xie2019dba}. Can we design an attack-adaptive defense strategy for robust federated learning?

In federated learning,the attackers may control a fraction of clients and manipulate the local data and the model updates to inject a backdoor or to degrade the global model's performance. For example, with the backdoor attack \cite{BagdasaryanVHES20}, the attacker can locally assign a `trash' label to the images of the automobiles manufactured by a certain brand and contaminate the global model. Therefore, it is important to defend the attacks for robust federated learning training. 

From the server's perspective, the only clue for defending the adversarial attacks is the model updates submitted from the clients, in contrast to the attackers' large flexibility in designing attacks. Also, the heterogeneous data distribution (non-identically distributed data) in federated learning make the problem more challenging.
Hence, the key of the defense is on designing a robust aggregation strategy for the model updates. In an aggregation strategy, we treat the model update as a vector, and we want to discard the corrupted update vectors from the attackers while keeping only the genuine update vectors from the benign clients.

Several aggregation rules, instead of FedAvg \cite{McMahanMRHA17}, have been proposed for defending the adversarial attacks. Classical robust estimators, such as Coordinate-wise median \cite{YinCRB18,chen2019distributed} and Geometric median (implemented as RFA in \cite{pillutla2019robust}) have been proposed but their performance degraded due to the heterogeneous data distribution in federated learning. Residual-based reweighing \cite{Fu19} extends the classical robust regression to the federated learning setting, but it has a low breakdown point. On the other hand, several similarity-based aggregation rules have been proposed. FoolsGold \cite{Fung18} asserts the similarity of attackers, Krum and its variants \cite{blanchard2017machine,mhamdi2018hidden} assert the similarity of the benign clients (in terms of Euclidean distance),  clustering-based approaches \cite{sattler2020byzantine,munoz2019byzantine} assert both with cosine similarity. Nevertheless, these similarity-based defenses can be bypassed by projecting the corrupted update vectors to the neighborhood of the genuine update vectors \cite{baruch2019little,BagdasaryanVHES20}. Recent works try to uncover more properties of the attacker. WeakDP \cite{sun2019can} tries to cancel the effect of the attacker by clipping and adding noise to the update vectors. However, the optimal size of noise is not well studied. In \cite{li2020learning}, their approach detects attackers by learning a variational autoencoder on randomly sampled coordinates of the unbiased model updates obtained in the centralized training setting. However, the optimal latent representation of the model updates is not well studied. Apart from their weakness, the former defense strategies may fail to detect edge-case backdoor \cite{wang2020attack}, where a very small region of the model updates are altered. It indicated the need of a tailor-made defense for challenging attacks.

In this work, we propose the first attack-adaptive aggregation mechanism for robust federated learning that learns to detect possible corrupted update vectors from challenging attacks. Our approach learns a low dimensional representation of the update vectors that allows detection of possible attacks.
Specifically, we train an attention \cite{vaswani2017attention} based neural network to explore the vulnerable regions of the model with respect to various attacks. We feed the update vectors to the attention module to obtain the alignment scores between the latent representations and reweigh the update vectors accordingly. We simulate the federated learning tasks under different attacks with the test set in the server and collect the update vectors to train our model in a self-supervised fashion. We show the approximation capacity of our neural network on similarity measures.

We compare our approach with existing aggregation rules on federated learning tasks: MNIST \cite{lecun1998gradient} classification, CIFAR-10 \cite{krizhevsky2009learning} classification, Tiny-ImageNet classification, and IMDb \cite{maas-EtAl:2011:ACL-HLT2011} sentiment analysis. Our approach outperforms prior work in defending model poisoning and backdoor attacks. Our approach can also generalize the defense to different datasets, numbers of clients, and attack parameters.

\section{Related work}

\subsection{Attacks on federated learning.}
Adversarial attacks can attack either the data or the model. \cite{blanchard2017machine} suggested the omniscient attack, which multiplies the update vector by a negative constant, can reverse the direction of gradient descent and degrade the model performance.  \cite{Fung18} suggested that the label flipping attack, which changes the label of a certain class, can already be an effective attack if there are no defenses.
\cite{BagdasaryanVHES20} showed that the backdoor attack
, which injects a certain pattern to the data and alters the label to the desired target, can mislead the global model while not affecting the standard accuracy. \cite{xie2019dba} proposed the distributed backdoor attack, which embeds similar but different patterns to the data, to bypass similarity-based defenses. Some works \cite{baruch2019little,BagdasaryanVHES20} show that certain defenses can be bypassed by projecting the corrupted update vectors to the neighborhood of the genuine ones. In our work, we will show the potential of a data-driven and attack-adaptive aggregation strategy in defending adversarial attacks.

\subsection{Robust federated learning.}
The defense on federated learning can be categorized in terms of robustness, privacy, and security.  In terms of robustness, several approaches defend adversarial attacks by designing aggregation rules as introduced in Section \ref{intro}. There are defenses relating to other aspects of federated learning. \cite{sun2019can} suggested that adding the differential privacy
can improve robustness against certain attacks. \cite{pillutla2019robust} proposed RFA, a secure implementation of the Geometric median aggregation rule. On the other hand, some approaches sacrifice certain level of privacy for robustness. Zeno \cite{xie2018zeno} audits the local models' accuracy on a test set in the server. \cite{wang2020attack} compares the update vectors with the unbiased model updates trained on the test set. 
In our work, we will focus on robustness, and we assume that the server has access to the update vectors and a test set that is disjoint from the local data. 



\section{Method}

\subsection{Formulation}
\paragraph{Federated learning.} 
In federated learning, the server distributes a global model $\vtheta_{global}$ to each of the $n$ clients. Then each client $i$ trains its local model $\vtheta_{client}^{(i)}$ with its own data and sends the update vector 
\[\vx_i=\vtheta_{client}^{(i)}-\vtheta_{global}\] 
back to the server. The server aggregates the set of update vectors $\{\vx_i\}$ by the aggregation strategy $g(\{\vx_i\})$ and updates the global model as 
\[
\vtheta_{global}\leftarrow \vtheta_{global}+g(\{\vx_i\}).
\]
The new global model is then distributed to the clients for the next round of training.

\paragraph{Robust aggregation strategy.}
Federated learning can be vulnerable to adversarial attacks. The attackers can attack their local data or the update vectors directly. One the other hand, the server knows only the clients' update vectors but not their local training data or even the number of samples trained locally. Hence, a robust aggregation strategy is key to defend the attacks. Let $\sD_{benign}$ to be a set of genuine update vectors from benign clients and  $\sD_{attack}$ to be a set of the corrupted update vectors from attackers. We denote the mean of only the genuine update vectors as the \textbf{robust mean}
\begin{equation}\label{robustmean}
    \vmu_{robust}=\sum_{i=1}^n \frac{ \1_{(\vx_i\in \sD_{benign})}}{\sum_{j=1}^n \1_{(\vx_j\in \sD_{benign})}}\vx_i,
\end{equation} where $\1_{(condition)}$ is the indicator function which evaluates to 1 if the condition is true and 0 otherwise. 
A robust aggregation strategy $g(\cdot)$ aims to approximate the robust mean $\vmu_{robust}$, i.e. solving the minimization
\begin{equation}\label{eq:minimization}
   \argmin_g \left\|g\left(\{\vx_i\}\right)-\vmu_{robust}\right\|.
\end{equation}
The difficulty of designing a robust aggregation strategy is that the attackers can evolve their attacks to bypass the current defense. Hence we are interested in an aggregation strategy that can readily be adapted to defend the challenging attacks. In Section \ref{datadriven}, we will propose a data-driven framework for attack-adaptive aggregation.

\subsection{Attack-adaptive aggregation}\label{datadriven}
This work provides a self-supervised way to detect attacks when aggregating update vectors in federated learning. We collect empirical data from federated learning tasks for training a data-driven model that detects corrupted update vectors. Our data-driven model is attack-adaptive because it can identify the vulnerable regions of the update vector with respect to different attacks. We simulate the federated learning tasks under different attack scenarios on the test set in the server. We collect the update vectors and their labels (corrupted or genuine). The data-driven model can then be trained with the update vectors as input and a loss function that encourages the prediction to agree with the label. With such a data-driven model, we can defend against the attacks missed out by the previous methods and refine our defense readily against new attacks. We may update our defense model incrementally and serve the new defense model as a `security patch'. From the clients' perspective, they could be informed of the anomaly in their local data or model, as well as which type of anomaly they are suspected of. The clients can then inspect the unintended contamination in their data accordingly.

To obtain such a data-driven model, we may parameterize the indicator function $\1_{(\vx_i\in \sD_{benign})}$ in Equation \ref{robustmean} by a neural network and retrain the neural network upon new attacks. However, the neural network would not work unless it can take arbitrary number of update vectors and arbitrary permutation of the clients since the order of the arrival of the update vectors is not fixed. Moreover, it should have the capacity to incorporate existing robust estimators, such as the Coordinate-wise median \cite{YinCRB18}, as prior knowledge. Hence we may parameterize instead the $p(\vx_i\in \sD_{benign}|\vq_t)$, which is the probability of $\vx_i$ being a genuine update vector from a benign client given a robust estimate $\vq_t$. Then we may get the next estimate by reweighing the update vector $\vx_i$ with the probability. Here we define $\vq$ to be our estimator of the robust mean $\vmu_{robust}$. We can obtain the estimator by the iteration:
\begin{equation}\label{iteration}
    \begin{split}
    \vq_0&=\mathrm{\textbf{med}}(\{\vx_i\}),\\
            \vq_{t+1}&=\sum_{i=1}^n\frac{ p(\vx_i\in \sD_{benign}|\vq_t) }{\sum_{j=1}^n p(\vx_j\in \sD_{benign}|\vq_t)}\vx_i,
    \end{split}
\end{equation}
where $\mathrm{\textbf{med}} (\{\vx_i\})$ is a function taking median coordinate-wisely on the update vectors $\{\vx_i\}$. The $\mathrm{\textbf{med}} (\{\vx_i\})$ serves as an initial guess and could be replaced with other robust estimators or simply the mean. In the iteration, the update vector $\vx_i$ is reweighed with the probability $p(\vx_i\in \sD_{benign}|\vq_t)$ which depends on $\vx_i$ and $\vq_t$. Such form of reweighing is very similar to the attention mechanism in neural network.


\paragraph{Attention}\label{aba}

\begin{figure*}[t]
\centering
\includegraphics[width=0.8\textwidth]{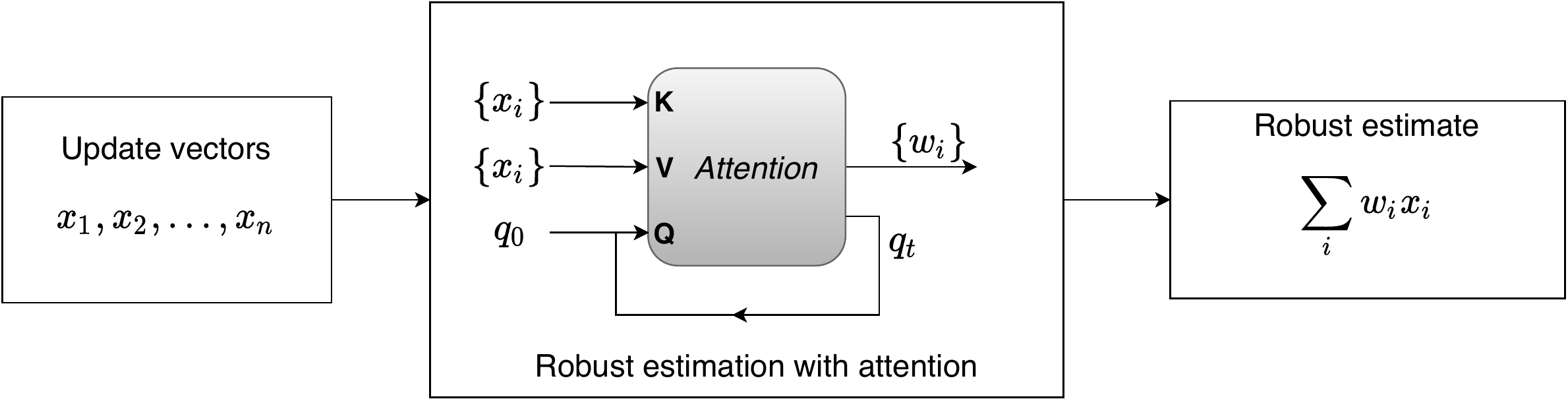}
\caption[The overview of our approach.]{The overview of our approach for estimating the robust mean $\vmu_{robust}$ of the update vectors.}
\label{diagram}
\end{figure*}

Our model is described in Algorithm \ref{algo}. Our model consists of multiple passes of an attention module. The update vectors' weights are updated in each pass, and a new estimate is obtained by reweighing the update vectors. The overview of our method is summarized in Figure \ref{diagram}.

\begin{algorithm}[tb]
   \caption{Attack-adaptive aggregation with attention}
   \label{algo}
\begin{algorithmic}
   \STATE {\bfseries Input:} update vectors $\{\vx_i\}$, hyperparameters $c,\varepsilon,T$
      \STATE {\bfseries Output:} robust estimate $\vq_T$, the weights of the update vectors $\{w_i\}$
   \STATE  $\vq_0=\mbox{\textbf{med}}(\{\vx_i\})$\;
   \FOR{$t=0$ {\bfseries to} $T-1$}
   \FOR{$i=1$ {\bfseries to} $n,$ \textbf{\textup{in parallel}}}
   \STATE $s_i=\frac{Q(\vq_t)\cdot K(\vx_i)}{\|Q(\vq_t)\|\|K(\vx_i)\|}$
    \STATE $w_i= { \exp{(cs_i)} }/{\sum_{j=1}^n \exp{(cs_j)}}$
    \STATE $w_i=w_i\cdot \1_{(w_i\geq \varepsilon/n)}$
   \ENDFOR
   \STATE $\vq_{t+1}=\sum_{i=1}^n w_i\vx_i$
   \ENDFOR
   \STATE \textbf{return} $\vq_{T}, \{w_i\}$
\end{algorithmic}
\end{algorithm}

In our approach, we use the attention mechanism for the parameterization of the likelihood $p(\vx_i\in \sD_{benign}|\vq_t)$. We encode the robust estimate $\vq_t$ by the query encoder $Q$, and the update vectors $\{\vx_i\}$ by the key encoder $K$ and the value encoder $V$. We fix the value encoder $V$ to be the identity and train the key encoder $K$ and query encoder $Q$ such that the alignment score 
\begin{equation}
    s_i=\frac{Q(\vq_t)\cdot K(\vx_i)}{\|Q(\vq_t)\|\|K(\vx_i)\|}
\end{equation}
is closed to $+1$ for genuine update vector $\vx_i\in \sD_{benign}$ and $-1$ for corrupted update vector $\vx_i\in\sD_{attack}$. 

The original version of attention in \cite{vaswani2017attention} does not fit our purpose of parameterizing the iteration in Equation \ref{iteration} since $e^{s_i}$ cannot cover the range from 0 to 1. Instead, we use the softmax with temperature \cite{guo2017calibration} and the overall expression in one pass of the attention module is
\begin{equation}\label{softmaxExpr}
\begin{split}
    \vq_{t+1}&=\frac{\sum_{i=1}^n e^{cs_i} \vx_i}{\sum_{i=1}^n e^{cs_i}}=\frac{\sum_{i=1}^n \left(e^{cs_i}/e^c\right) \vx_i}{\sum_{i=1}^n \left(e^{cs_i}/e^c\right)},
\end{split}
\end{equation}
where the scale factor $c=1/\tau$ is the inverse of the temperature $\tau$.

Here we can observe that the form in Equation \ref{softmaxExpr} is very similar to that in the Equation \ref{iteration}. The only difference is that the probability term $p(\vx_i\in \sD_{benign}|\vq_t)$ in  Equation \ref{iteration} is replaced with  $e^{cs_i}/e^c$. The term $e^{cs_i}/e^c$ has a range very close to $[0,1]$ for a large $c$. Also, it contains the information of the update vectors $\vx_i$ and the last estimate $\vq_t$. Therefore, we can see that $e^{cs_i}/e^c$ is a suitable representation of the probability term $p(\vx_i\in \sD_{benign}|\vq_t)$. In another perspective, the corrupted update vectors are assigned a lower weight when we have a larger $c$.

In our algorithm, we further add a truncation step with the threshold $\varepsilon/n$  after we compute the softmax values. It is done to eliminate the effect of any corrupted update vectors with a potentially large magnitude. The truncation step
\begin{equation}
    w_i\leftarrow w_i\cdot \1_{(w_i\geq \varepsilon/n)}
\end{equation}
zeroes out the attention weight if it is smaller than the threshold $\varepsilon/n$. It is necessary because the exponent $e^{-c}$ can never reach $0$, and it can be problematic if we have a corrupted update vector with an extremely large magnitude, for example, $e^{2c}$.

\section{Implementation}
\subsection{Dimensionality reduction}
One difficulty of training our model is that the dimension of the update vector is very large compared to the number of clients $n$. The model may overfit to the irrelevant regions. In fact, since we are concerning the relative deviation of the update vectors, we can operate on the low-rank approximation of the set of update vectors. By performing PCA and assuming the update vectors are already centered (since the update vectors represent changes), we get a low dimension representation of the update vectors. Moreover, the vulnerable regions of the model in a federated learning task may reside in multiple layers. Hence, we perform PCA for each layer instead of performing it once for the whole update vectors. We keep all of the $n$ principal components in each layer and the layer-wise PCA corresponds to a rotation in each layer.

After dimensionality reduction, we apply our model in Algorithm \ref{algo} to estimate the robust mean of the projected update vectors and their corresponding weights. The robust mean of the original update vectors can then be estimated by reweighing with the same weights. 


A limitation of using PCA directly is that the attacker may hide its attack in multiple directions. For instance, the attacker in \cite{bhagoji2019analyzing} adds a $l_2$-regularization on the distance to the previous benign updates. However, such $l_2$-regularized attack may not be stealthy in our case, where PCA is performed on each layer, and deviation at any layer may be flagged by our defense.
Suppose the tolerable $l_2$-deviation is at most $\varepsilon$ at each of the $L$ layers and, as a result, the total deviation is at most $\sqrt{L}\varepsilon$. In this case, hiding the attack in a $l_2$-ball of size $\sqrt{L}\varepsilon$ is sufficient to bypass the plain PCA, but a size of $\varepsilon$ is required to bypass our layer-wise version. Hence, the attacker needs to strengthen the $l_2$-regularization by a factor $\sqrt{L}$. Moreover, our attention module further suggests the vulnerable regions of the projected update vector. To hide the attack, the attacker needs to regularize further the cosine distance $\mathrm{cos}(\vv(\vx_{benign}),\vv(\vx_{attack}))$, where $\vv$ is a projection to the vulnerable regions and $\vv$ may not be known to the attacker. Our approach restricts the forgery at each layer and the vulnerable regions of the update vector. Hence, the attacker gets a worse trade-off between the stealthiness and effectiveness of its attack.

\subsection{Training}
We only need to consider the query encoder $Q$ and the key encoder $K$ to train our model. In our work, both encoders are 2-layer multi-layer perceptrons with ReLU activations. We perform the forward pass as described in Algorithm \ref{algo}. We obtain the predicted estimate and compare it with the ground truth robust mean $\vmu_{robust}$ described in Equation \ref{robustmean}. We use the $L_1$ loss and the Adam optimizer for the backpropagation. We train our model for 500 epochs with $T=5$. For each attack, we run the federated learning tasks three times on the test set in the server to collect update vectors. Update vectors from two of the runs are used for training our model. The remaining run is served for validation. 


\subsection{Hyperparameter search}
In our model, there are two major hyperparameters: $c$ and $\varepsilon$. We perform a hyperparameter sweep to find a combination that yields a high validation accuracy on predicting attackers.
We found that a set of moderate values around $c=10, \varepsilon=0.5$ is a good choice, and we use these values for our implementation. For the other hyperparameters, we found that they are also not sensitive. For instance, $T=1$ is sufficient to reject the attackers, further passes to the attention module make the weights of the benign clients more uniform.

\section{Experiments}
\subsection{Experimental setup}

We compare our aggregation strategy with 6 prior works: FedAvg \cite{McMahanMRHA17},  Coordinate-wise median \cite{YinCRB18}, RFA \cite{pillutla2019robust}, Krum \cite{blanchard2017machine}, FoolsGold \cite{Fung18}, and Residual-based reweighing \cite{Fu19}. We evaluate the performance of the aggregation strategies on four federated learning tasks under different attacks. To simulate a heterogeneous data distribution, we divide each dataset into disjoint partitions with the Dirichlet distribution with hyperparameter 0.9 as in \cite{BagdasaryanVHES20}. Different from \cite{hsu2019measuring}, we do not require the clients to have the same number of samples when generating the partitions.
In each round, the clients train their local models on their data for one epoch. Then all clients, including possible attackers, are selected for the aggregation. Some prior works require hyperparameters. For Krum, we set $m=\floor{\frac{n}{2}}-2$. For FoolsGold, we set $\kappa=1$. For Residual-based reweighing, we set $\lambda=2, \delta = 0.1$.

\subsection{Tasks}\label{task}
\paragraph{MNIST classification} In this task, we use a LeNet \cite{lecun1998gradient} model with 10 clients. We evaluate the federated learning tasks under three types of attacks. (No attack) It simulates federated learning on heterogeneous data. (Omniscient) The attackers negate their update vectors by multiplying them by $-1$. (Backdoor) The attackers embed a pixel pattern to 50\% of their image samples and alter their label to digit `2'. We run the task for 30 communication rounds.

\paragraph{CIFAR-10 classification}\label{cifar-task}
In this task, we use a ResNet-18 \cite{he2016deep} model with 10 clients. We evaluate the federated learning tasks under three types of attacks: no attack, omniscient, backdoor. We run the task for 30 communication rounds.

\paragraph{Tiny-ImageNet classification} We use the same model and the same attack scenarios as in CIFAR-10 classification. We run the task for 45 communication rounds.

\paragraph{IMDb sentiment analysis}
 In this task, we use a Gated recurrent unit \cite{cho2014learning} with FastText embedding \cite{joulin2017bag} with 10 clients. We evaluate the federated learning tasks under three types of attacks. (No attack) It simulates the federated learning on heterogeneous data distribution. (Label flipping) The attackers swap the label of class `Positive' and class `Negative' in their local data. (Omniscient) The attackers negate their update vectors. We run the task for 10 communication rounds.


\subsection{Metrics}
We evaluate the performance of the aggregation strategies on the standard accuracy and the attack success rate. The accuracy (\textbf{ACC}) refers to the global model's standard accuracy on the test set. The attack success rate (\textbf{ASR}) is an evaluation of federated learning training against backdoor attacks. It measures how many backdoor-injected samples are classified as the target label of the attacker. If a backdoor-attacked sample is predicted to be the target label, we consider the attack on this sample is successful. 

Our metrics are defined by
\begin{equation}
    \begin{split}
\mbox{\textbf{ACC}}&=\frac{\mbox{\# correct predictions}}{\mbox{\# samples}},\\
\mbox{\textbf{ASR}}&=\frac{\mbox{\# successfully attacked samples}}{\mbox{\# attacked samples}}.
\end{split}
\end{equation}
The higher the accuracy, the better the aggregation strategy defends the attacks from interfering with the federated learning task. The lower the attack success rate, the better the aggregation strategy defends the backdoor attacks. An ideal aggregation strategy can achieve 100\% accuracy and has the attack success rate as low as the fraction of attacked samples from the target class. We report the accuracy and the attack success rate in the final communication round. We take the average over three runs.

\subsection{Evaluation}
\label{evaluation}
\paragraph{Visual tasks}
\begin{figure}[t]
    \centering
    \includegraphics[width=0.45\textwidth]{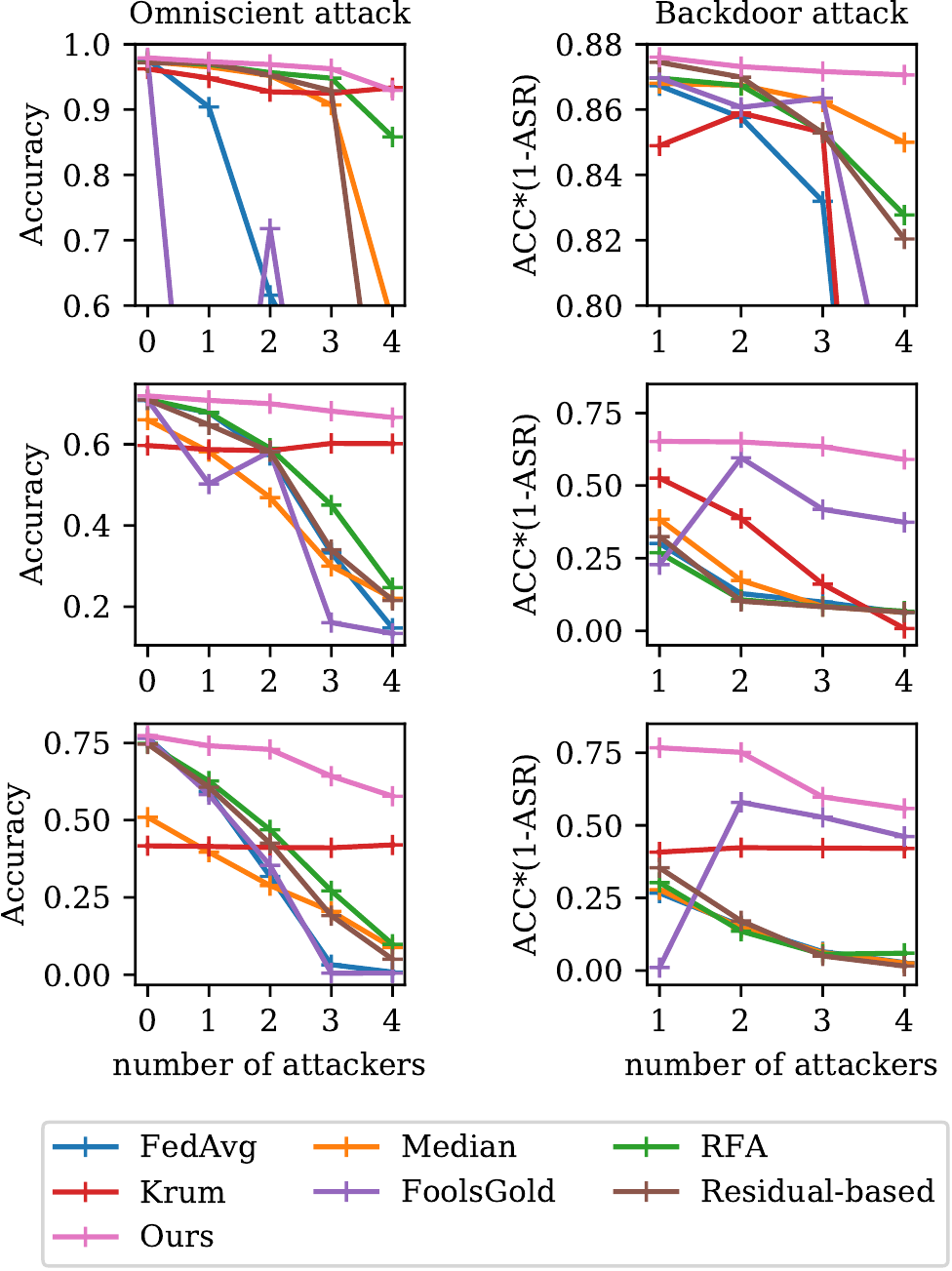}
    \caption[The performance of aggregation strategies in the visual tasks.]{ The performance of aggregation strategies in the MNIST (top), CIFAR(middle), ImageNet(bottom) classification task. 
    The `no attack' scenario is combined with the omniscient attack in the left plots.   } 
    \label{result_visualTask}
\end{figure}

The results of the MNIST, CIFAR, Tiny-ImageNet tasks are summarized in Figure \ref{result_visualTask}. The detailed results for backdoor attack can be found in the Appendix \ref{apdix:backdoor_results}. Under omniscient attack, most of the defense strategies failed when there were more attackers. It could be attributed to the curse of dimensionality, where negation of the vector does not alter the pairwise similarity much. In contrast, our approach was more resilient in all three visual tasks across different numbers of attackers since our approach operated on the projected update vectors where the negation became obvious. However, when there were 4 attackers in the most complicated Tiny-ImageNet task, the convergence was not good, even our approach outperformed others. The reason could be that the number of benign clients was too low to learn an effective global model in this complicated task. Nevertheless, our approach performed better than other approaches and had a good convergence when there was a moderate number of attackers.

Regarding the backdoor attack, our approach had the highest \textbf{ACC}*(1-\textbf{ASR}) score in all visual tasks. It indicated that our approach had both a high global model accuracy and a low attack success rate. In the simplest MNIST task with abundant data, the backdoor was 'forgotten' when the global model became more mature over the communication rounds. Nevertheless, our approach removed the effect of the backdoor more effectively. In the CIFAR-10 and the Tiny-ImageNet task, the backdoor was not 'forgotten' since the ResNet-18 had more capacity to learn both the main task and the backdoor. Most approaches failed when there were 4 attackers and have $>90\%$ \textbf{ASR}. In these tasks, the benign update vectors had a larger variance, and the corrupted update vectors from the backdoor attackers were relatively similar. It explained the higher resilience of FoolsGold when there were more attackers. Nevertheless, FoolsGold still had a $>40\%$ \textbf{ASR} while our approach had a $<15\%$ \textbf{ASR} when there were 4 attackers. It indicated that our approach could better identify the vulnerable regions of the update vectors. On the other hand, Krum learned the backdoor in CIFAR-10 task and did not learn both the backdoor and the main task well in the Tiny-ImageNet task since it aggregated only a small fraction of clients. In contrast, our approach aggregated most of the benign clients and achieved a high global accuracy and a low attack sucess rate.

\paragraph{Textual task}
\begin{figure}[t]
    \centering
    \includegraphics[width=0.45\textwidth]{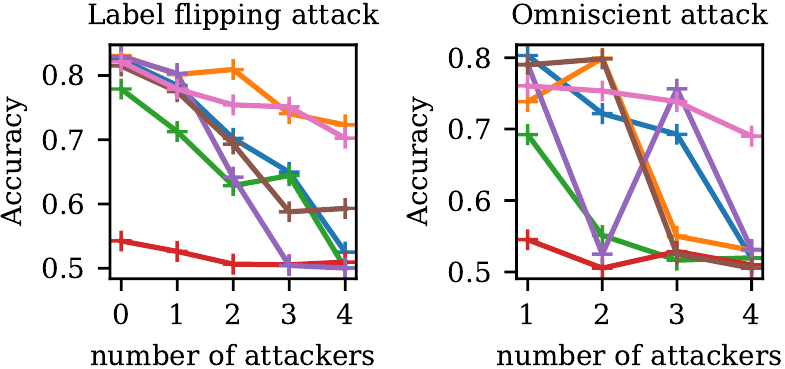}
    \caption[The performance of aggregation strategies in the textual task.]{The performance of aggregation strategies in the IMDb sentiment analysis task. We use the same legend in Figure \ref{result_visualTask}. Our approach is in pink.}
    \label{imdb-result}
\end{figure}
The Figure \ref{imdb-result} shows the results on the IMDb sentiment analsysis task. most approaches degraded quickly with the number of attackers. Some of them were even worse than FedAvg. Seemingly, acquiring a benign client was more important than discarding an attacker in this task. Nevertheless, our approach managed to maintain a relatively high accuracy against the attacks. When there were 4 omniscient attackers, our approach was the only one that worked.

\subsection{Ablation study}

\begin{table}[t]
\begin{center}\resizebox{0.45\textwidth}{!}{

\begin{tabular}{lrrrrr}
\toprule
{}  & \multicolumn{5}{l}{Attack Sucess Rate} \\
\# of attackers &                      1 &      2 &      3 &      4 & Average \\
\midrule
FedAvg         &              57.37 &  81.87 &  86.03 &  91.02 &   79.07 \\
MLP         &              54.35 &  79.92 &  86.25 &  89.60 &   77.53 \\
w/o $c$            &                15.21 &  57.42 &  84.12 &  88.90 &  61.41\\
w/o $\varepsilon$           &       9.78 &  10.39 &  11.19 &  32.30 &  15.92\\
\midrule
Ours        &           \textbf{6.78} &   \textbf{5.64} &   \textbf{6.29} &  \textbf{13.03} &  \textbf{  7.94 }\\
\bottomrule
\end{tabular}}
\end{center}
\caption{Ablation study on the effect of removing attention (MLP), scaled softmax (w/o $c$), or the truncation step (w/o $\varepsilon$).}
\label{tab:cifar_ablation}
\end{table}

We performed an ablation study on various components of our model. We evaluated the effectiveness of the defenses on the CIFAR-10 task under the backdoor attack.
The results are summarized in Table \ref{tab:cifar_ablation}. The scale factor $c$ in our softmax played an important role in defending attacks. The threshold factor $\varepsilon$ controlled the trade-offs between robustness and convergence. Without attention, a plain multi-layer perceptron overfitted a certain permutation of the clients and could not distinguish the attackers when they arrived in a different order. Our attention-based model avoided this problem since it is permutation invariant.

\subsection{Transferability of defense}
\begin{figure}
    \centering
    \includegraphics[width=0.48\textwidth]{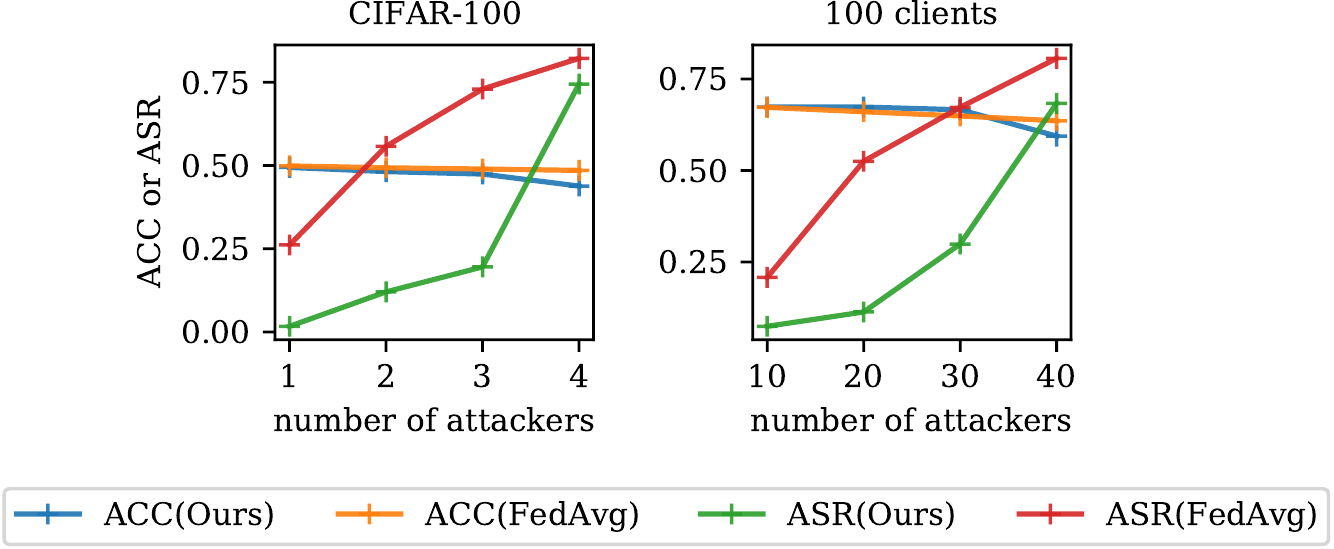}
    \caption[The performance of defense against backdoor attack across different scenarios.]{Performance of defense against backdoor attack across different scenarios. (Left) Transfering the defense to CIFAR-100 task.  (Right) Transfering the defense on 100 clients in CIFAR-10 task.  }
    \label{fig:transfer}
\end{figure}

\begin{figure}
    \centering
    \includegraphics[width=0.48\textwidth]{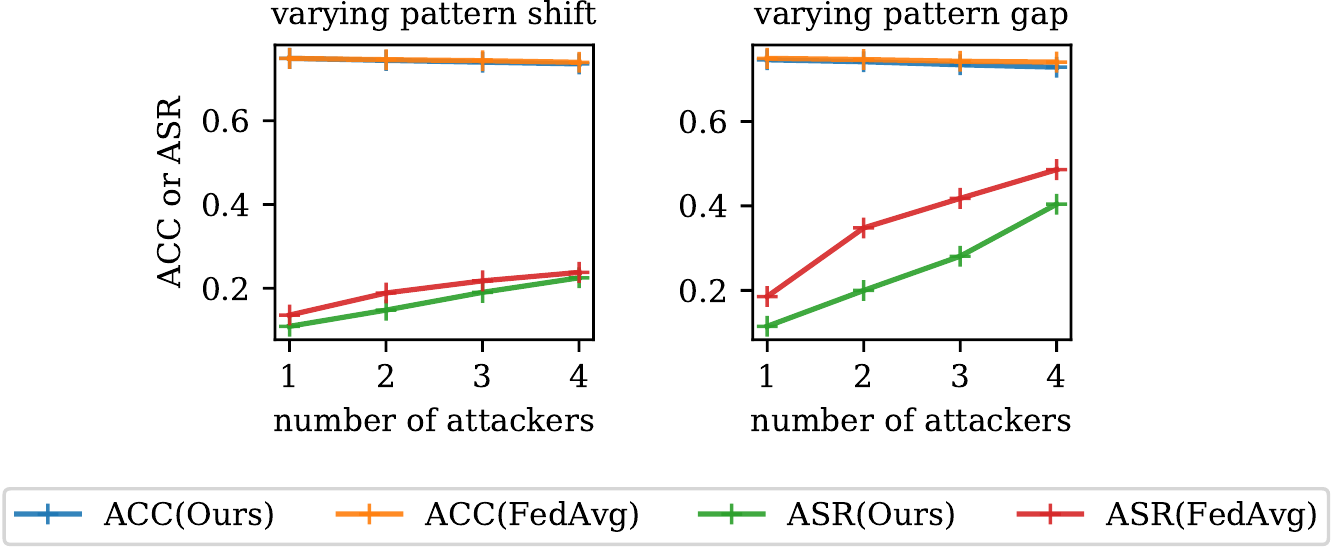}
    \caption[The performance of defense against backdoor attack with different backdoor patterns.]{Performance of defense against backdoor attack with different backdoor patterns. We varied the values of the parameters of the backdoor patterns and reported the average result w.r.t. each parameter. }
    \label{fig:pattern_avg}
\end{figure}

As a practical data-driven aggregation mechanism, it is important to know how far our defense can be generalized to unseen scenarios. We study the transferability of our defense. Specifically, we trained our model on the CIFAR-10 task under backdoor attacks with a fixed backdoor pattern. Then we evaluated the defenses' performance under different scenarios: 1.) classification tasks on CIFAR-100 instead of CIFAR-10, 2.) 100 clients instead of 10 clients,  and 3.) different backdoor patterns. The results is shown in Figure \ref{fig:transfer}. Our approach generalized the defense better when there was a lower fraction of attackers. On the other hand, Figure \ref{fig:pattern_avg} showed that our approach could generalize the defense to unseen backdoor patterns. In summary, our defense can be generalized to unseen attack scenarios when there are a lower fraction of attackers. 
It meets our expectations since our approach learns the similarity measure based on the attacks' traits of a set of plausible attack scenarios. With a new attack scenario, it may share part of the vulnerable regions and the deviation in these regions remains detectable when there are a lower fraction of attackers. On the other hand, our model is less confident to accuse a client when there are a high fraction of attackers. To address the issue, we need to provide additional supervision on the backdoor patterns that we want to defend.


\section{Conclusion}

In this work, we presented a novel approach for robust federated learning using a deep neural network as the aggregation function. To the best of our knowledge,  our aggregation strategy is the first one that is attack-adaptive and learns to defend against various attacks in a data-driven fashion. The attention mechanism in our designed network is effective in propagating contextual information to detect malicious attackers. We further demonstrate the transferability of our defense. We hope our attack-adaptive aggregation paradigm can inspire more work in this direction. 
Our source code is publicly available on \href{https://github.com/cpwan/Attack-Adaptive-Aggregation}{https://github.com/cpwan/Attack-Adaptive-Aggregation}.

\newpage
\bibliographystyle{named}
\bibliography{main}

\appendix
\include{append}

\end{document}

%% file: math_commands.tex
\usepackage{amsmath,amsfonts,bm}

\def\eqref#1{equation~\ref{#1}}

\def\floor#1{\lfloor #1 \rfloor}
\def\1{\bm{1}}

\def\vmu{{\bm{\mu}}}
\def\vtheta{{\bm{\theta}}}

\def\vq{{\bm{q}}}

\def\vv{{\bm{v}}}

\def\vx{{\bm{x}}}

\DeclareMathAlphabet{\mathsfit}{\encodingdefault}{\sfdefault}{m}{sl}
\SetMathAlphabet{\mathsfit}{bold}{\encodingdefault}{\sfdefault}{bx}{n}

\def\sD{{\mathbb{D}}}

\DeclareMathOperator*{\argmin}{arg\,min}

%% file: append.tex
\section{Illustrations of data and model}
\subsection{Generating heterogeneous data distribution}\label{ext:data}

\begin{figure}[ht]
\centering
    \includegraphics[width=0.45\textwidth]{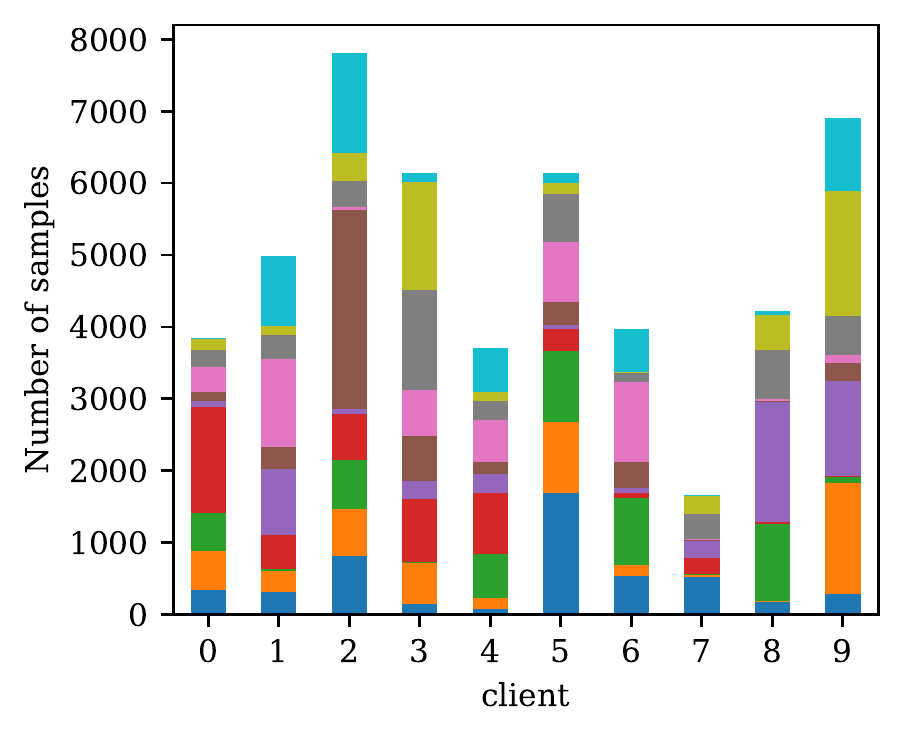}
    \includegraphics[width=0.45\textwidth]{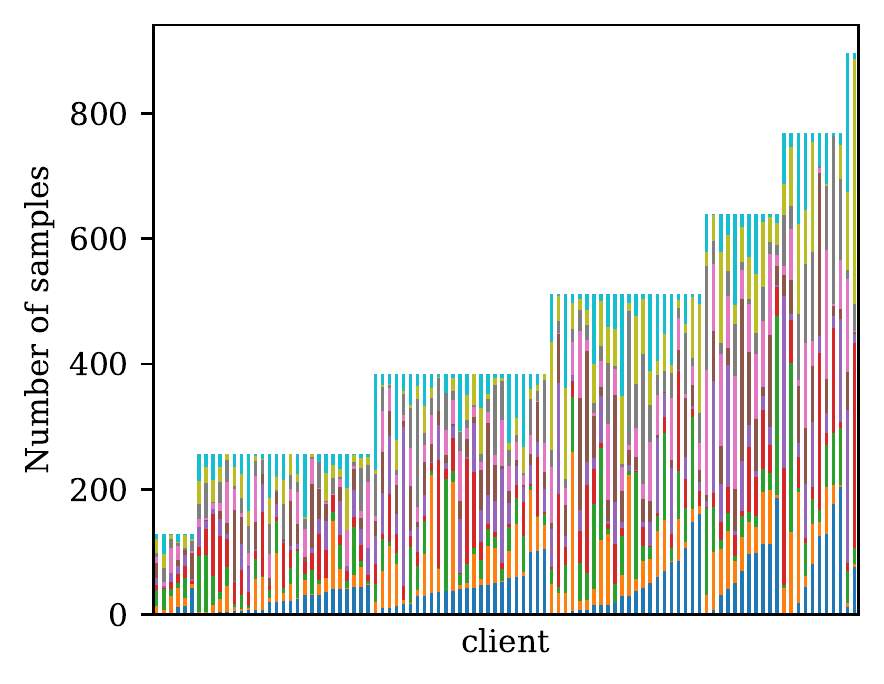}
    \caption[An illustration of the data distribution across clients.]{ An illustration of the data distribution across clients. Each color corresponds to a class. (Left) 10 clients. (Right) 100 clients. The clients are sorted solely for visualization purposes. The clients appear in random order when we run the experiments. We used a batch size of 128 in the local training. Hence the total number of samples in each client is a multiple of 128. }
    \label{partition}
\end{figure}

In our experiments, we generated heterogeneous data distribution with the Dirichlet distribution. An instance of the generated data distribution is shown in Figure \ref{partition}. We used the Dirichlet distribution with the concentration parameter $\boldsymbol{\alpha}=(0.9,0.9,\hdots,0.9)$ to generate the fraction of samples to be drawn for each client. We performed the partition for each label and assigned the drawn samples to the clients. It was different from some previous works, which forced the clients to have the same number of data and drawn the data samples with replacement. In our data distribution generation, the data were divided into \textbf{disjoint} partitions with varying sizes. In this way, the federated learning experiments could be run on different data distributions. It allowed us to collect update vectors from diverse federated learning settings, which provided higher-quality data for training our model.

\subsection{Demonstrations on synthetic data}\label{sec:demo}

\begin{figure}[h]
    \centering
    \subfigure[An instance of synthetic data]{
         \centering
         \includegraphics[width=0.25\textwidth]{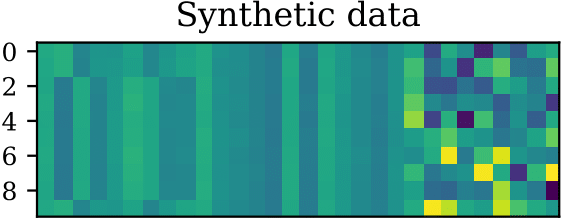}
         \label{fig:example}
     }
     \subfigure[PCA on synthetic data]{
         \centering
         \includegraphics[width=0.25\textwidth]{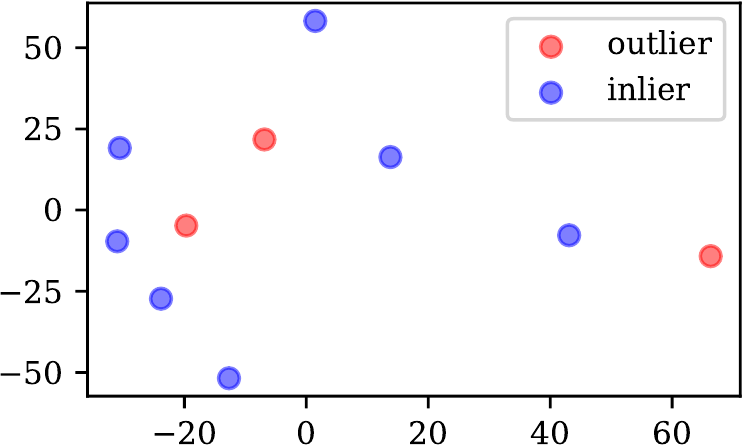}
         \label{fig:examplePCA}
    }
   \subfigure[Weights assigned by our model]{
         \centering
         \includegraphics[width=0.20\textwidth]{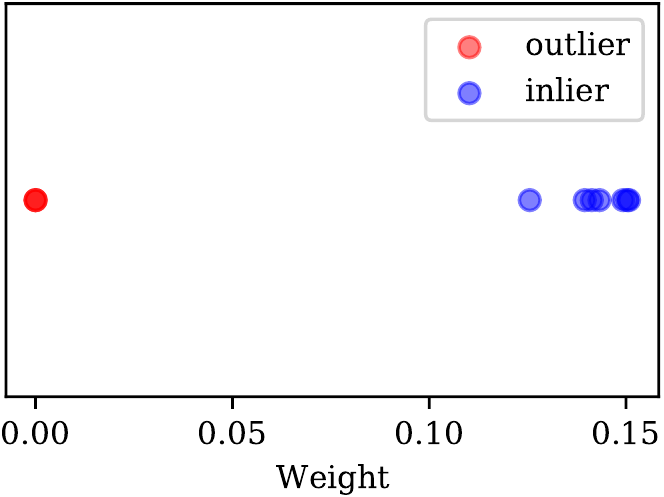}
         \label{fig:exampleAnalysis}
   }
     \caption[An illustration of the synthetic data.]{Illustration of outlier detection on synthetic data. The samples 0, 1, 9 are the outliers. }
\end{figure}

We will demonstrate how our model works for robust estimation on synthetic data. Each instance of the synthetic data is a set of 10 samples drawn from a 30-dimensional multivariate normal distribution. The last one-third of features consist of noise, while the outliers had their first one-third of features modified. We generated a training set of size 2048. Figure \ref{fig:example} illustrated an instance of the synthetic data in the validation set, in which the sample 0, 1, 9 are the outliers.  Figure \ref{fig:examplePCA} showed no apparent distinctions between inliers and outliers under PCA. We trained our model on the 2048 instances of synthetic data. The results in Figure \ref{fig:exampleAnalysis} illustrated that our model successfully distinguished the outliers. 

\begin{figure}[h]
    \centering
    \includegraphics[width=0.3\textwidth]{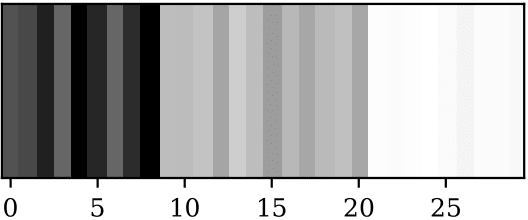}
    \caption[ The feature importance of the synthetic data.]{The feature importance on the 30 features of the synthetic data, estimated from our model (the darker, the more important).}
    \label{fig:impt}
\end{figure}

Apart from estimating each individual set of samples, we also estimate the feature importance by measuring the magnitude of the weight $W$ in the first layer of our key encoder for the $i$-th feature:
\begin{equation}\label{importance_score}
    \mathrm{imptz}(i)=\sum_{j=0}^h \left|W_{i,j}\right|,
\end{equation}
where $h$ is the dimension of the hidden layer. Figure \ref{fig:impt} shows that our model indeed captured the features that distinguish outliers from inliers. We may employ similar visualization to explore which part of the model is more vulnerable to adversarial attacks in federated learning. 

\newpage

\section{More experiments}

\subsection{Capturing the traits of attacks}\label{ext:traits}
\begin{figure}[!ht]
    \centering
\subfigure[FedAvg 
]{
    \includegraphics[width=0.23\textwidth]{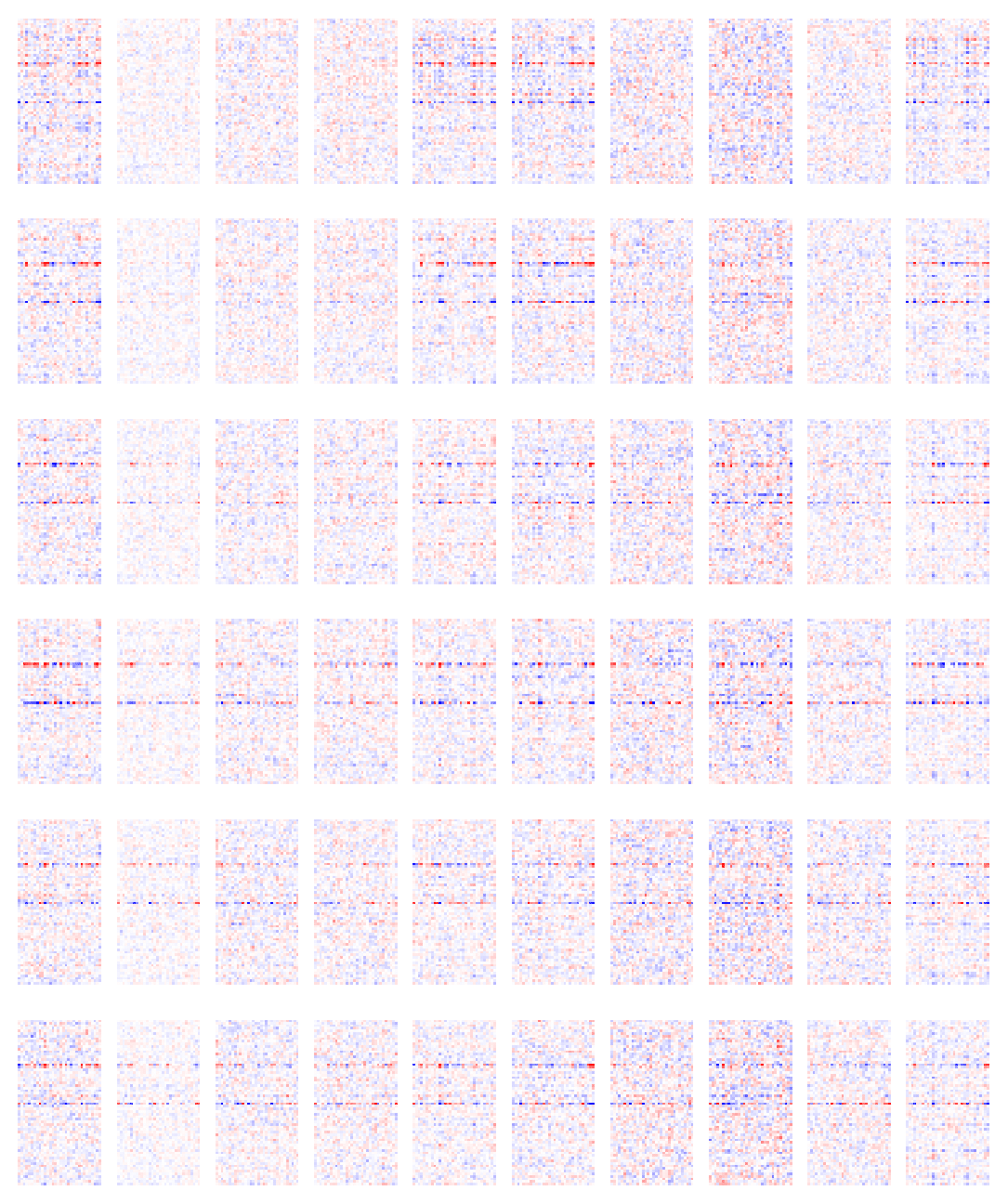}}
    \subfigure[Ours
    ]{
\includegraphics[width=0.23\textwidth]{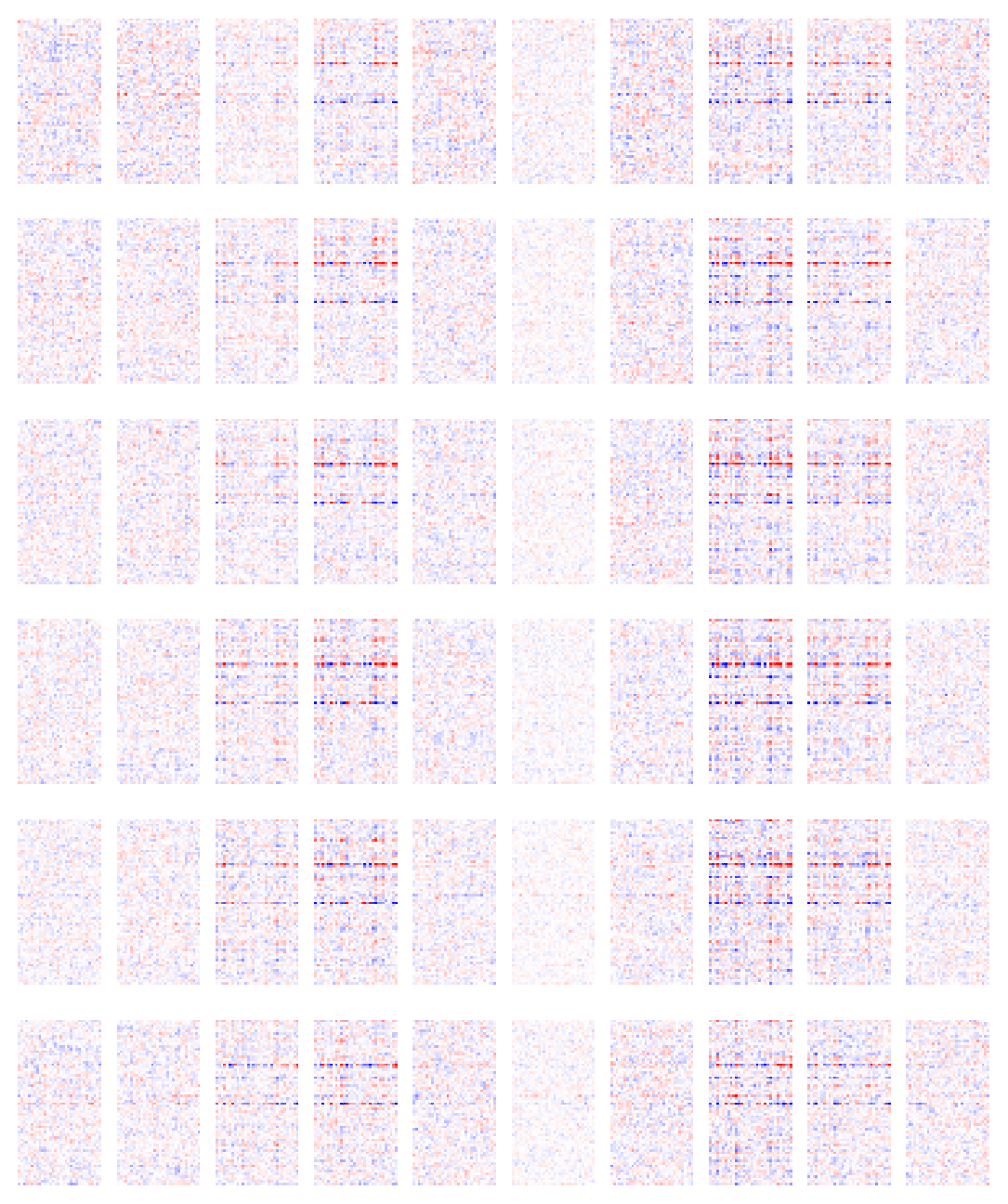}}
    \caption[The snapshots of the update vectors under different attack scenarios.]{The \texttt{layer3.0.downsample.0.weight} layer in the (0,2,4,6,8,10)-th communication round in the CIFAR-10 task under backdoor attack with FedAvg (Left) and our attack-adaptive aggregation (Right). The columns correspond to the clients. The blue color means positive, the red color means negative. The darker is the color, the larger is the magnetuide. The model parameters are reshaped for illustration purposes. For the task with FedAvg, the client 0, 4, 5, 9 is the attacker. For the task with our attack-adaptive aggregation, the client 2, 3, 7, 8 is the attacker. }
    \label{fig:analysis_net}
\end{figure}

We may analyze the feature importance on the attack-adaptive aggregation model we trained similarly as in section \ref{sec:demo} . We will use the same definition from Equation \ref{importance_score}. In this section, we will analyze our attack-adaptive aggregation model. The model were trained to defend against backdoor attack in the CIFAR-10 tasks with 10 clients. We summed up the feature importance for each layer (with weight and bias separately counted). The most prominent layers are
\begin{itemize}
    \item \texttt{layer3.0.downsample.1.weight},
    \item  \texttt{layer3.0.downsample.0.weight},
    \item  \texttt{layer4.1.bn2.bias},
    \item  \texttt{layer4.0.downsample.1.weight}
\end{itemize}
The result is reasonable. Since the backdoor attacker attempts to recognize the backdoor pattern, the attacker's local model needs to extract features in its residual blocks. Although the receptive field is sufficient to cover the pattern (with a shape of 3 by 7) in the earlier blocks, the later blocks seem responsible for learning the pattern. Besides, it seems that the downsampling layers are more sensitive to the backdoor pattern. It may be due to that the attacker favors some channels during downsampling. Indeed, when we look at the \texttt{layer3.0.downsample.0.weight} layer of the update vectors in Figure \ref{fig:analysis_net}, the two horizontal strips of the attackers indicate that the corresponding two channels are given a higher weight during downsampling.

Based on the literature on neural network architecture, we may already have some ideas on the neural network's vulnerability. However, it is not immediately trivial for one to decide which layers may contain the attack's traits. On the other hand, our attack-adaptive aggregation model can readily discover the attack's traits and offer a defense at the same time. It is illustrated in Figure \ref{fig:analysis_net} that the traits of the attack we found indeed affected the training. For FedAvg, the update vectors of the benign clients were progressively contaminated by the two horizontal strips. It implied that the attackers had successfully led the gradient direction to a malicious objective. In contrast, with our attack-adaptive aggregation, the benign clients could deliver uncontaminated update vectors and continued the global training with an appropriate gradient direction. The result confirms the ability of our attack-adaptive aggregation to identify the traits of attack.

In summary, our approach serves as a defense strategy and provides a way to analyze the adversarial attack's traits. In contrast, some previous approaches did not consider the contributions of different regions of the update vector. In this regard, our approach is an interpretable defense strategy.

\subsection{Transferability of defense across backdoor patterns}\label{ext:transfer}
\begin{figure}
    \centering
    \includegraphics[width=0.14\textwidth]{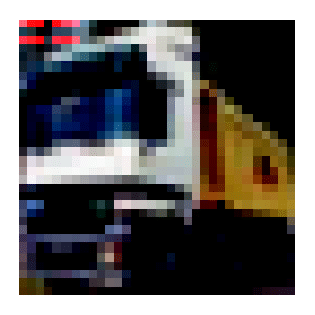}
    \includegraphics[width=0.33\textwidth]{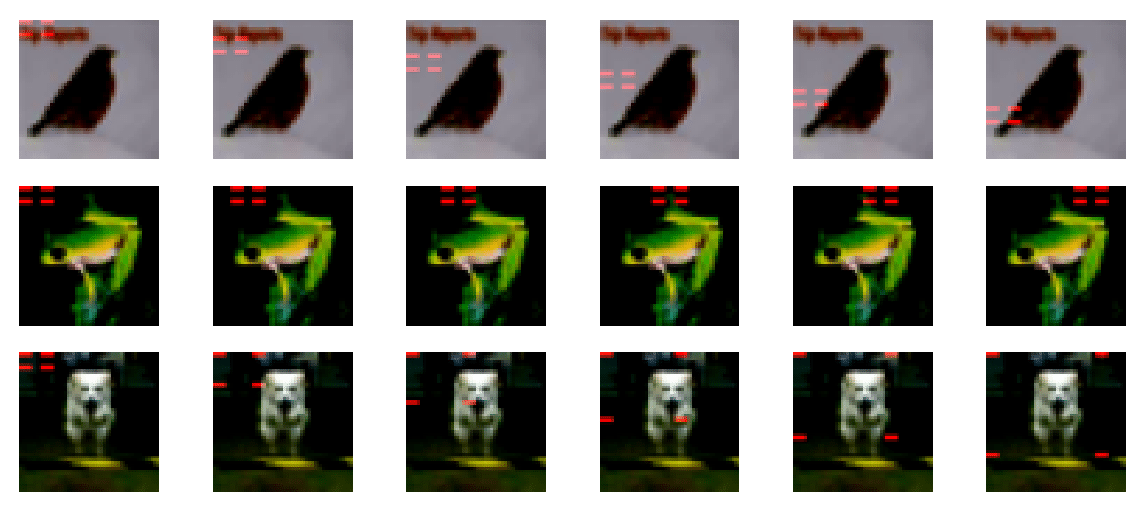}
    \caption[Illustration of the \textit{shift} and \textit{gap} parameters of the backdoor pattern.]{Backdoor patterns. (Left) The red pixel pattern is the backdoor pattern used in training our attack-adaptive aggregation. (Right) From top to bottom, each row demonstrates how the backdoor pattern looks like in different $\textit{shift}_y$, $\textit{shift}_x$, and $\textit{gap}$. The  $\textit{shift}_y$ and $\textit{shift}_x$ parameters control the shift of the pattern along  and vertical and horizontal direction respectively. The $\textit{gap}$ parameter controls how many gap to be injected between the bars in additional to the default 1 pixel gap. }
    \label{fig:pattern_demo}
\end{figure}

In the main text, we summarized the average results of transferring the defense to different backdoor patterns. In this section, we will give the results in detail. During the training of our attack-adaptive aggregation, we used the pixel pattern shown in Figure \ref{fig:pattern_demo}. The backdoor pattern occupies a 3 (pixels) by 7 (pixels) region in the top-left corner of the image. It consists of four 1 by 3 horizontal bars arranged in two columns, with a 1-pixel gap. The backdoor attacker modified the red channel of these 12 pixels to the largest intensity. In each experiment on transferability, we letted the attackers to use a backdoor pattern with different $\textit{shift}_y,\textit{shift}_x$, and $\textit{gap}$ parameters. The  $\textit{shift}_y$ and $\textit{shift}_x$ parameters control the shift of the pattern along  and vertical and horizontal direction respectively. The $\textit{gap}$ parameter controls how wide the gap is between the bars in addition to the default 1-pixel gap. In the evaluation, we varied only one parameter and kept the other parameter fixed. In additional, for experiments on $\textit{shift}_y$ and $\textit{shift}_x$, we set $\textit{gap}=1$ to introduce two-pixels gaps between the bars in the backdoor pattern (instead of only one). It was done to ensure that the new backdoor pattern is different from the one we used in training. 

\begin{figure}[hp!]
    \centering
    \subfigure[Varying $\textit{shift}_y$.]{
    \includegraphics[width=0.45\textwidth]{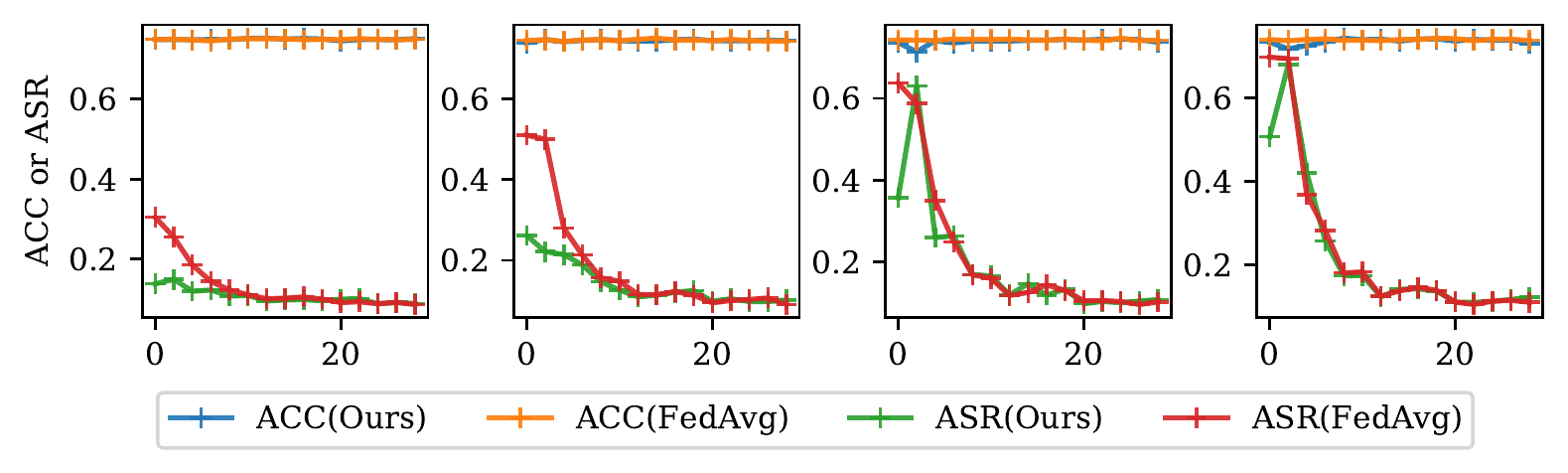}
    }
    \subfigure[Varying $\textit{shift}_x$.]{
    \includegraphics[width=0.45\textwidth]{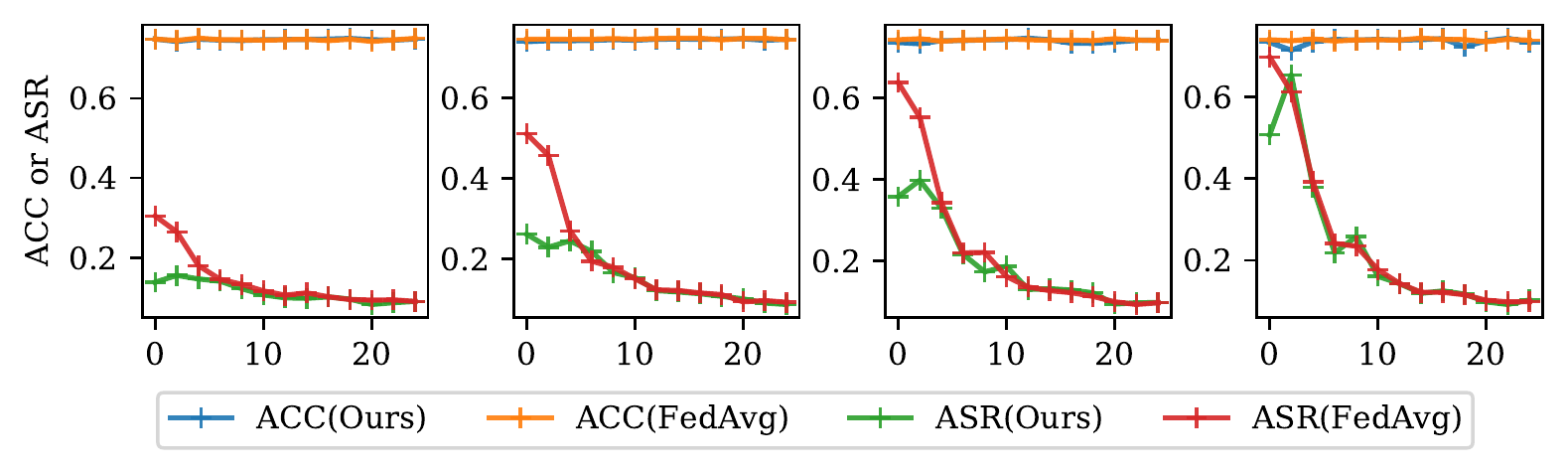}
    }
    \subfigure[Varying $\textit{gap}$.]{
    \includegraphics[width=0.45\textwidth]{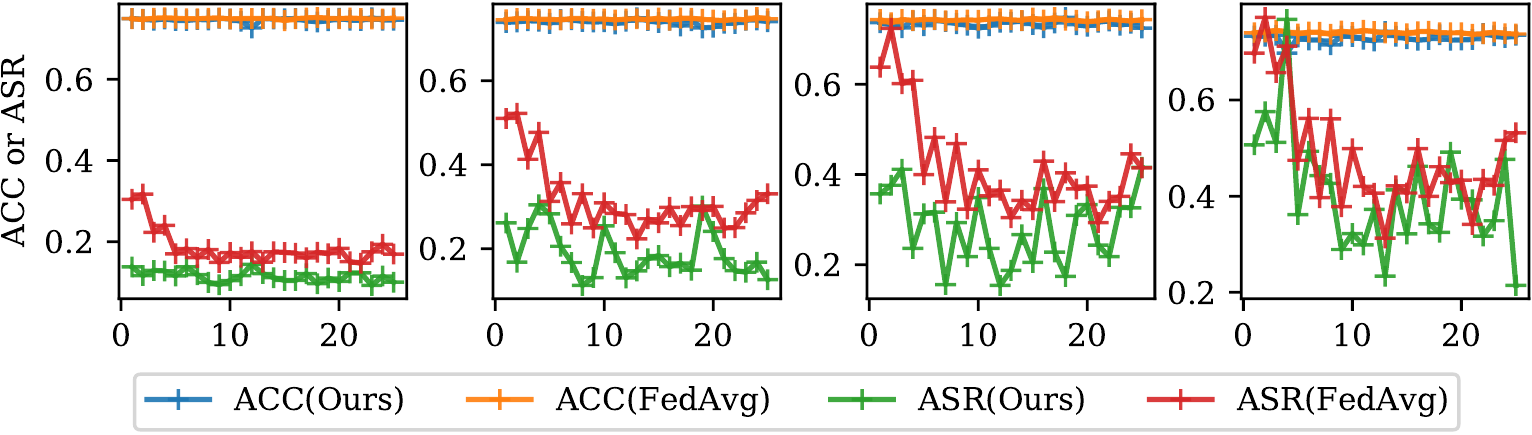}
    }
    \caption[The effects of backdoor patterns on the transferability of our approach.]{Effects of backdoor patterns on the transferability of our attack-adaptive aggregation. From left to right, the plots show the performance of the defense strategies under 1 to 4 attackers. \textbf{ACC} stands for the standard accuracy (higher the better). \textbf{ASR} stands for the attack success rate (lower the better).}
    \label{fig:change_transfer}
\end{figure}

The effects of $\textit{shift}_y,\textit{shift}_x$, $\textit{gap}$ parameters are shown in Figure \ref{fig:change_transfer}. For the $\textit{shift}_y,\textit{shift}_x$ parameters, the backdoor attack was generally weaker when the backdoor pattern was further away from the top-left corner. It agrees to the previous research 
that the local model may `forget' the backdoor pattern in the middle of the image. On the other hand, it is shown that our attack-adaptive aggregation was able to generalize the defense across different backdoor pattern parameters when there were 1 to 2 attackers. Nevertheless, our attack-adaptive aggregation did not generalize well enough to defend 3 to 4 attackers. Since the attack's traits in the update vectors could be different when a different backdoor pattern is used, our attack-adaptive aggregation may treat some of the new traits of the different attack as a  naturally occurred variance. For instance, if the new backdoor pattern is on the top-right corner instead, then the backdoor attack may leave a different trait in the update vector. If there is only one attacker, such a trait may still be noticeable by our model. However, if there are multiple attackers, our model cannot decide whether such a trait results from an attack or images containing an object in the top-right corner. As a result, our attack-adaptive aggregation could not be confident enough to accuse the attackers. To address the issue, we need to train a new attack-adaptive aggregation model with update vectors collected under different attack parameters.

\newpage
\subsection{Detailed results of the backdoor attacks}\label{apdix:backdoor_results}
Table \ref{tab1},\ref{tab2},\ref{tab3} show the detailed results of the visual tasks under backdoor attacks described in Section \ref{evaluation}.
\begin{table*}[!tpbh]
\begin{center}
\resizebox{0.8\textwidth}{!}{
\begin{tabular}{lrrrrrrrrrr}
\toprule
{} & \multicolumn{5}{l}{Accuracy} & \multicolumn{5}{l}{Attack Sucess Rate} \\
\# of attackers &        1 &      2 &      3 &      4 & Average &                  1 &      2 &      3 &      4 & Average \\
\midrule
FedAvg         &    97.09 &  96.76 &  96.43 &  96.02 &   96.58 &              10.67 &  11.37 &  13.73 &  38.84 &   18.65 \\
Median         &    96.77 &  96.75 &  96.46 &  95.61 &   96.40 &              10.30 &  10.34 &  10.60 &  11.10 &   10.58 \\
RFA            &    97.08 &  96.88 &  96.39 &  96.37 &   96.68 &              10.42 &  10.47 &  11.51 &  14.11 &   11.63 \\
Krum           &    94.99 &  95.46 &  94.98 &  95.26 &   95.17 &              10.63 &  10.02 &  10.19 &  42.66 &   18.38 \\
FoolsGold      &    97.23 &  96.47 &  96.85 &  96.25 &   96.70 &              10.55 &  10.78 &  10.84 &  22.34 &   13.63 \\
Residual-based &    97.39 &  97.09 &  96.34 &  95.55 &   96.59 &              10.21 &  10.40 &  11.48 &  14.14 &   11.56 \\
\midrule
Ours           &    97.40 &  96.90 &  96.83 &  96.79 &   96.98&              \textbf{10.06} &  \textbf{ 9.89} &   \textbf{9.98} &  \textbf{10.05} &   \textbf{10.00} \\
\bottomrule
\end{tabular}}
\end{center}
\caption{Performance on MNIST task under backdoor attack.}
\label{tab1}
\begin{center}
\resizebox{0.8\textwidth}{!}{
\begin{tabular}{lrrrrrrrrrr}
\toprule
{} & \multicolumn{5}{l}{Accuracy} & \multicolumn{5}{l}{Attack Sucess Rate} \\
\# of attackers &        1 &      2 &      3 &      4 & Average &                  1 &      2 &      3 &      4 & Average \\
\midrule
FedAvg         &    70.51 &  70.61 &  70.85 &  69.13 &   70.28 &              57.37 &  81.87 &  86.03 &  91.02 &   79.07 \\
Median         &    64.31 &  66.60 &  65.39 &  64.82 &   65.28 &              40.33 &  73.96 &  87.13 &  90.19 &   72.90 \\
RFA            &    71.03 &  70.42 &  70.10 &  69.42 &   70.24 &              62.17 &  84.83 &  88.07 &  90.28 &   81.34 \\
Krum           &    58.31 &  60.76 &  57.23 &  58.71 &   58.75 &               9.78 &  36.34 &  71.94 &  98.64 &   54.18 \\
FoolsGold      &    70.52 &  69.54 &  68.81 &  67.69 &   69.14 &              67.75 &  14.30 &  39.24 &  44.77 &   41.52 \\
Residual-based &    70.72 &  70.04 &  69.61 &  69.40 &   69.94 &              54.10 &  85.45 &  88.09 &  90.78 &   79.60 \\
\midrule
Ours           &    69.97 &  68.93 &  67.70 &  67.86 &   68.62 &               \textbf{6.78} &   \textbf{5.64} &   \textbf{6.29} &  \textbf{13.03} &  \textbf{  7.94 }\\
\bottomrule
\end{tabular}}
\end{center}
\caption{Performance on CIFAR-10 task under backdoor attack.}
\label{tab2}
\begin{center}
\resizebox{0.8\textwidth}{!}{
\begin{tabular}{lrrrrrrrrrr}
\toprule
{} & \multicolumn{5}{l}{Accuracy} & \multicolumn{5}{l}{Attack Sucess Rate} \\
\# of attackers &        1 &      2 &      3 &      4 & Average &                  1 &      2 &      3 &      4 & Average \\
\midrule
FedAvg         &    62.79 &  53.43 &  41.56 &  32.75 &   47.63 &              57.50 &  71.21 &  84.45 &  91.60 &   76.19 \\
Median         &    41.73 &  32.99 &  25.00 &  17.88 &   29.40 &              33.30 &  55.35 &  75.44 &  85.39 &   62.37 \\
RFA            &    64.17 &  49.68 &  41.37 &  29.24 &   46.12 &              52.81 &  72.77 &  86.46 &  79.62 &   72.91 \\
Krum           &    40.93 &  42.50 &  42.42 &  42.24 &   42.02 &              \textbf{ 0.38} &  \textbf{ 0.46} & \textbf{  0.57} &   \textbf{0.48} &    \textbf{0.47} \\
FoolsGold      &    16.02 &  70.03 &  67.82 &  60.91 &   53.70 &              93.73 &  17.30 &  22.18 &  24.28 &   39.37 \\
Residual-based &    63.91 &  49.96 &  40.50 &  31.64 &   46.50 &              44.67 &  65.83 &  87.64 &  95.10 &   73.31 \\
\midrule
Ours           &    77.06 &  75.54 &  60.18 &  59.24 &  { 68.01} &              { 0.45} &   {0.53} &   {0.67} &  { 5.80 }&    {1.86} \\
\bottomrule
\end{tabular}
}
\end{center}
\caption{Performance on ImageNet task under backdoor attack. Note that Krum has a poor accuracy even it achieves the lowest attack success rate. Our approach achieves both a low attack success rate and a high accuracy. }
\label{tab3}
\end{table*}

\subsection{Training with multiple attacks}

We evaluated the effect of training our defense on multiple types of attacks. We compared the performance of our defense against the attacks when the defense was trained on 1.) only the backdoor attack scenarios, 2.) only the omniscient attack scenarios, or 3.) both the backdoor attack and the omniscient attack scenarios. The results of our defenses against backdoor attack is summarized in Table \ref{cifar-backdoorm}. Interestingly, even we trained our defense only on the omniscient attack scenarios but not the backdoor attack scenarios, the defense can still defend the backdoor attack. This is reasonable because the omniscient attack negates the update vector and every coordinate in the update vector could be considered vulnerable under the omniscient attack. Therefore, our attention module takes every coordinate of the update vector into account, including the coordinates involved with the backdoor attack. So, the anomaly in these coordinates could still be detected. On the other hand, when we train the defense with the backdoor attack, our attention module learned the vulnerable regions better and had a better defense when there were more attackers. Similarly, we can defend the omniscient attack even we trained our defense only on the backdoor attack scenarios, as shown in Table \ref{cifar-omnim}. This is again due to the overlapping vulnerable regions with respect to the two attacks. When there were 4 attackers, our defense was again stronger if we trained our defense on the same attack. 

We observed that we can defend a different type of attack if the attack shares the vulnerable region with the attack scenario that we trained on. However, when we train our defense on multiple attacks, our defense may not perform as good as training alone on a single attack, especially in the cases of higher fraction of attackers. It may due to that our attention module learned a suboptimal vulnerable regions when multiple attacks were involved in the training. When our defense tries to increase the detection rate of an attack, it may also raise the chance of false alarm in another attack scenario. Therefore, our defense may become more conservative and do not work as good when there are higher fraction of attackers. Nevertheless, the performance of our defense trained on multiple attacks was still competitive when compared with other aggregation strategies.

\begin{table*}[!htbp]

\begin{center}
\resizebox{0.8\textwidth}{!}{
\begin{tabular}{lrrrrrrrrrr}
\toprule
{} & \multicolumn{5}{l}{Accuracy} & \multicolumn{5}{l}{Attack Sucess Rate} \\
\# of attackers &        1 &      2 &      3 &      4 & Average &                  1 &      2 &      3 &      4 & Average \\
\midrule
Backdoor (B)         &  69.97 &  68.93 &  67.70 &  67.86 &   68.62 &               \textbf{6.78} &   \textbf{5.64} &   \textbf{6.29} &  \textbf{13.03} &  \textbf{  7.94 }\\
Omniscient (O)         &    70.06 &  68.92 &  69.02 &  66.36 &   68.59 &              8.70 &  10.12 &  9.44 &  34.23 &   15.62 \\
B+O           &    69.85 &  69.13 &  68.23 &  67.87 &   68.77 &               {7.56} &   {8.16} &   {13.05} &  {38.02} &  {  16.70 }\\
\bottomrule
\end{tabular}}

\end{center}
\caption{Performance of our approach on CIFAR-10 task under \textbf{backdoor attack} when trained with different attack scenarios.}\label{cifar-backdoorm}
\end{table*}

\begin{table*}[!htbp]

\begin{center}
\resizebox{0.48\textwidth}{!}{
\begin{tabular}{lrrrrr}
\toprule
{} & \multicolumn{5}{l}{Accuracy}\\
\# of attackers &        1 &      2 &      3 &      4 & Average \\
\midrule
Backdoor (B)         &  70.46 &  \textbf{68.47} &  67.41 &  59.78 &   66.53              \\
Omniscient (O)         &    70.79 &  67.97 &  \textbf{68.14} &  \textbf{66.59} &   \textbf{68.87}    \\
B+O           &    \textbf{70.85} &  68.46 &  {66.69} & 44.73 &   62.68 \\
\bottomrule
\end{tabular}}

\end{center}
\caption{Performance of our approach on CIFAR-10 task under \textbf{omniscient attack} when trained with different attack scenarios.}\label{cifar-omnim}
\end{table*}

\newpage
\section{Theoretical Analysis}

\subsection{Universal approximation property}\label{ext:UATkernel}
We use the attention module in our attack-adaptive aggregation. The query encoder $Q$ and the key encoder $K$ in the attention module are 2-layer multi-layer perceptrons with ReLU activation. We will show in Theorem \ref{uatTheo} that given large enough hidden units and large enough latent space in the last layer of $Q$ and $K$, the dot product $Q(\cdot)\cdot K(\cdot)$ can approximate any similarity measure and the alignment score function $\frac{Q(\cdot)\cdot K(\cdot)}{\|Q(\cdot)\|\|K(\cdot)\|}$ is a projection of such a similarity measure to $[-1,1]$.

\begin{theorem}\label{uatTheo}
Let $f_1, f_2: [-M,M]^{d_j}\to [-M,M]^{D'}$ be continuous functions, $h:[-M,M]^{D'} \to \mathbb{R}$ be a symmetric, continuous positive definite kernel function, $\sigma(\cdot)$ be ReLU. Then, for arbitrary $\varepsilon'>0,$ by specifying sufficiently large $D, T\in \mathbb{N}$, there exist $\boldsymbol{A} \in \mathbb{R}^{D \times T}, \boldsymbol{B} \in \mathbb{R}^{T \times d_j}, \boldsymbol{c} \in \mathbb{R}^{T}$
such that
$$
\left|h\left(f_{1}(\boldsymbol{x}), f_2\left(\boldsymbol{x}^{\prime}\right)\right)-\left\langle f_{1}^{\psi}(\boldsymbol{x}), f_{2}^{\psi}\left(\boldsymbol{x}^{\prime}\right)\right\rangle\right|<\varepsilon'
$$
for all $\left(\boldsymbol{x}, \boldsymbol{x}^{\prime}\right) \in[-M, M]^{d_1+d_2}$
where $ f_{i}^{\psi}\left(\boldsymbol{x}\right)=\boldsymbol{A} \boldsymbol{\sigma}\left(\boldsymbol{B} \boldsymbol{x}+\boldsymbol{c}\right)$
are two-layer neural networks with $T$ hidden units, $D$ dimension output layer and $\boldsymbol{\sigma}(\boldsymbol{x})$ is element-wise $\sigma(\cdot)$ function.
\end{theorem}

The Theorem \ref{uatTheo} is a special case of the Theorem 5.1 in \cite{OkunoHS18}. We apply their result for our theorem. The theorem implies that the dot product of two neural networks can approximate any similarity measure. Suppose we feed the robust mean to the query encoder $Q$ and the sets of update vectors to the key encoder $K$. In that case, the theorem implies that the encoders have the approximation ability such that the alignment score $\frac{Q(\cdot)\cdot K(\cdot)}{\|Q(\cdot)\|\|K(\cdot)\|}$ is close to $+1$ for genuine update vectors and is close to $-1$ for corrupted update vectors.

\subsection{Robust mean estimation as an optimization problem}
In each step $t$ of our algorithm, 

\begin{equation*}
\resizebox{\linewidth}{!}{
    $\begin{aligned}
&\left\|g\left(\{\vx_i\}\right)-\vmu_{robust}\right\|=\left\| \vq_t-\vmu_{robust}\right\|\\
&=\left\| \sum_{i=1}^n tr\left(\frac{ \left(e^{cs_i}/e^c\right) }{\sum_{k=1}^n \left(e^{cs_k}/e^c\right)}\right)\vx_i-\sum_{i=1}^n\frac{ \1_{(\vx_i\in \sD_{benign})}}{\sum_{k=1}^n \1_{(\vx_k\in \sD_{benign})}}\vx_i\right\|
    \end{aligned}$}
\end{equation*}
where $tr(*)$ is a truncation function that yields zero if $*<\varepsilon/n$

When training our model, we used the $L_1$ loss and  $T=5$, as stated in Section 4.2. \textbf{This is exactly minimizing $\left\|g\left(\{\vx_i\}\right)-\vmu_{robust}\right\|$ w.r.t. the minimizer $s_i$ at the last time step $T=5$}. Hence, the quality of $s_i$ decides the quality of our algorithm for the minimization. In our algorithm, $s_i$ is the result of the dot product of the encoders. That is why we need Theorem 1 to show that the minimizer $s_i$ can be sufficiently optimal.

\subsection{Error bound of the robust mean estimation} \label{ext:err_bound}

Here, we attempt to give an error bound for estimating the robust mean with our attack-adaptive aggregation model. Suppose $\vmu_{robust}\in \mathbb{R}^k$ is the robust estimate. We want a similarity measure $h'(k,q)$ such that for $\vq'$ in the neighborhood of $\vmu_{robust}$, $h'(\vx_i,\vq')= 1$ for the genuine update vector $\vx_i$ and $h'(\vx_j,\vq')= -1$ for the corrupted update vector $\vx_j$. In the rest of this section, $\|*\|$ stands for the $l_1$ norm.

We denote $s(k,q)=\frac{K(k) \cdot Q({q})}{\|K(k)\|\|Q({q}\|)}$ for the attention score between $k$ and $q$. Theorem \ref{uatTheo} suggests that we can train a neural network such that $s(k,q)$ approximates $h'(k,q)$. That is, for $\vq_{t-1}$ in $\delta$-neighborhood of $\vmu_{robust}$ relative to the update vectors where $\frac{\|\vq_{t-1}-\vmu_{robust}\|}{ \max_l{\left(\|\vx_l\|\right)}}\leq \delta$,
\begin{equation}\label{eq:condition}
\begin{split}
    \|1-s(\vx_i,\vq_{t-1})\| < \varepsilon'&\quad \forall \vx_i \in \sD_{benign}\\
        \|-1-s(\vx_j,\vq_{t-1})\| < \varepsilon'&\quad \forall \vx_j \in \sD_{attack}.
\end{split}
\end{equation}
for some small $\varepsilon'>0$. We will use this definition in Lemma \ref{lem1}.

\begin{lemma}\label{lem1}
Suppose $s(k,q)$ approximates $h'(k,q)$ for $\vq_{t-1}$ in the $\delta$-neighborhood of $\vmu_{robust}$ relative to the update vectors with an error bounded by $\varepsilon'$. Then the next estimate $\vq_{t}$ in our algorithm would have an error bounded by 
\begin{equation}\label{prove1}
    \max{\left(e^{c\varepsilon'}-1, \frac{n}{n-m}e^{c\varepsilon'}e^{-2c}\right)} \max_{l,w_l\geq \varepsilon/n}{\left(\|\vx_l\|\right)},
\end{equation}
where $c$ is the scale factor, $\varepsilon$ is the threshold factor defined in our algorithm, and $w_l$ is the attention score of update vector $x_l$.

Moreover, the error bound improves by a rate $\gamma$ if 
\begin{equation}\label{prove2}
\begin{split}
        \varepsilon'&\leq \min{\left(\frac{1}{c}\ln(\gamma\delta+1),2-\frac{1}{c}\ln{\left(\frac{\gamma^{-1}\delta^{-1}-m/n}{1-m/n}\right)}\right)}\\&=O(\gamma\delta).
\end{split}
\end{equation}
\end{lemma}

\begin{proof}
Since the attention score is normalized, we have 
\begin{equation*}
\begin{split}
    s(\vx_i,\vq_{t-1}) > 1-\varepsilon' &\quad \forall \vx_i \in \sD_{benign}\\
        s(\vx_j,\vq_{t-1})< -1+\varepsilon'&\quad \forall \vx_j \in \sD_{attack}.
\end{split}
\end{equation*}

Suppose we have $m$ attackers and $n-m$ benign clients, then

\begin{equation*}    
\begin{split}
    (n-m)e^{c(1-\varepsilon')} + m e^{-c} &< \sum_k e^{cs(x_k,\vq_{t-1})} \\&< (n-m)e^{c}+m e^{-c(1-\varepsilon')}.
\end{split}
\end{equation*}

Hence, $\forall \vx_i \in \sD_{benign}$, 

\begin{equation*}
\begin{split}
\frac{e^c}{(n-m)e^{c(1-\varepsilon')} + m e^{-c}}
         &> \frac{e^{cs(\vx_i,\vq_{t-1})}}{\sum_k e^{cs(\vx_k,\vq_{t-1})}} 
         \\&> \frac{e^{c(1-\varepsilon')}}{(n-m)e^{c}+m e^{-c(1-\varepsilon')}},\\
 \frac{1}{(n-m)e^{-c\varepsilon'} + m e^{-2c}}
 &> \frac{e^{cs(\vx_i,\vq_{t-1})}}{\sum_k e^{cs(\vx_k,\vq_{t-1})}} 
 \\&> \frac{e^{-c\varepsilon'}}{(n-m)e^{-c\varepsilon'} + m e^{-2c}}
\end{split}   
\end{equation*}
and thus,
\begin{equation}\label{eq:bound_benign}
\begin{split}
    &\left\|w_i-  \frac{\1_{(\vx_i\in \sD_{benign})}}{\sum_{k=1}^n \1_{(\vx_k\in \sD_{benign})}}\right\|\\
    &=\left\|\frac{e^{cs(\vx_i,\vq_{t-1})}}{\sum_k e^{cs(\vx_k,\vq_{t-1})}}-  \frac{\1_{(\vx_i\in \sD_{benign})}}{\sum_{k=1}^n \1_{(\vx_k\in \sD_{benign})}}\right\|\\
    &=\left\|\frac{e^{cs(\vx_i,\vq_{t-1})}}{\sum_k e^{cs(\vx_k,\vq_{t-1})}}-  \frac{1}{n-m}\right\|\\
    &\leq \frac{\max{(1-e^{-c\varepsilon'}-\frac{m}{n-m}e^{-2c}, e^{-2c} )}}{(n-m)e^{-c\varepsilon'} + m e^{-2c}} \\
    &= \max{(\frac{e^{2c}-e^{c(2-\varepsilon')}-\frac{m}{n-m} }{(n-m)e^{c(2-\varepsilon')} + m},\frac{ 1 }{(n-m)e^{c(2-\varepsilon')} + m})}
\end{split}
\end{equation}

On the other hand,
$\forall \vx_j \in \sD_{attack}$, 

\begin{equation*}
\begin{split}
\frac{e^{c(-1+\varepsilon')}}{(n-m)e^{c(1-\varepsilon')} + m e^{-c}}
         &> \frac{e^{cs(\vx_j,\vq_{t-1})}}{\sum_k e^{cs(\vx_k,\vq_{t-1})}} \\
         &> \frac{e^{c(-1)}}{(n-m)e^{c}+m e^{-c(1-\varepsilon')}},\\
\frac{e^{c\varepsilon'}}{(n-m)e^{2c} + m e^{c\varepsilon'}}
 &> \frac{e^{cs(\vx_j,\vq_{t-1})}}{\sum_k e^{cs(\vx_k,\vq_{t-1})}} \\
 &> \frac{1}{(n-m)e^{2c} + m e^{c\varepsilon'}}
\end{split}   
\end{equation*}
 and thus,
 \begin{equation}\label{eq:bound_attacker}
\begin{split}
    &\left\|w_j-  \frac{\1_{(\vx_j\in \sD_{benign})}}{\sum_{k=1}^n \1_{(\vx_k\in \sD_{benign})}}\right\|\\ &=\left\|\frac{e^{cs(\vx_j,\vq_{t-1})}}{\sum_k e^{cs(\vx_k,\vq_{t-1})}}-  \frac{\1_{(\vx_j\in \sD_{benign})}}{\sum_{k=1}^n \1_{(\vx_k\in \sD_{benign})}}\right\|\\
    &=\left\|\frac{e^{cs(\vx_ij,\vq_{t-1})}}{\sum_k e^{cs(\vx_ik,\vq_{t-1})}}-  \frac{0}{n-m}\right\|\\
    &\leq \frac{e^{c\varepsilon'}}{(n-m)e^{2c} + m e^{c\varepsilon'}}\\
    &=\frac{1}{(n-m)e^{c(2-\varepsilon')} + m}
    \end{split}
\end{equation}

Combining the results for the benign clients and the attackers, we have the error bound for estimating the robust mean:

\begin{equation}
\resizebox{\linewidth}{!}{$
 \begin{aligned}
     &\left\|\vq_t-\vmu_{robust}\right\| \\&= 
     \left\| \sum_l w_l\cdot \1_{(w_i\geq \varepsilon/n)}\vx_l-\vmu_{robust}\right\|
     \\&=
     \left\|\sum_l tr\left(\frac{ e^{cs(\vx_il,\vq_{t-1})}}{\sum_k e^{cs(\vx_ik,\vq_{t-1})}}\right)\vx_l-  \sum_l\frac{\1_{(\vx_i\in \sD_{benign})}}{\sum_{k=1}^n \1_{(\vx_k\in \sD_{benign})}}\vx_l\right\| \\ 
     & \leq  \sum_i \left\|\frac{e^{cs(\vx_ii,\vq_{t-1})}}{\sum_k e^{cs(\vx_ik,\vq_{t-1})}}-  \frac{\1_{(\vx_i\in \sD_{benign})}}{\sum_{k=1}^n \1_{(\vx_k\in \sD_{benign})}} \right\| \left\|\vx_i\right\| \\&\quad + \sum_{j, w_j\geq \varepsilon/n} \left\|\frac{e^{cs(\vx_ij,\vq_{t-1})}}{\sum_k e^{cs(\vx_ik,\vq_{t-1})}}-  \frac{\1_{(\vx_i\in \sD_{benign})}}{\sum_{k=1}^n \1_{(\vx_k\in \sD_{benign})}}\right\| \left\|\vx_j\right\|\\
     & \leq 
     \left((n-m)\max{(\frac{e^{2c}-e^{c(2-\varepsilon')}-\frac{m}{n-m} }{(n-m)e^{c(2-\varepsilon')} + m},\frac{ 1 }{(n-m)e^{c(2-\varepsilon')} + m})} \right.\\&\quad+\left.
     \frac{m}{(n-m)e^{c(2-\varepsilon')} + m}
     \right)
     \max_{l,w_l\geq \varepsilon/n}{\left(\|\vx_l\|\right)}\\
     &=
     \frac{\max{\left((n-m)e^{c(2-\varepsilon')} (e^{c\varepsilon'}-1), n \right) }}{(n-m)e^{c(2-\varepsilon')} + m}
     \max_{l,w_l\geq \varepsilon/n}{\left(\|\vx_l\|\right)}\\
     & \leq \max{\left(e^{c\varepsilon'}-1, \frac{n}{(n-m)e^{c(2-\varepsilon')} + m}\right)} \max_{l,w_l\geq \varepsilon/n}{\left(\|\vx_l\|\right)}
     \end{aligned}$}
\end{equation}
The first equality is due to the design of the algorithm. In the second equality, we denote $tr(*)$ to be the truncation function that yields zero if $*<\varepsilon/n$. The second term of the second equality is due to the definition of robust mean $\mu_{robust}$. The third line is obtained by applying triangle inequality and splitting the terms for $\vx_i$ from benign clients and $\vx_j$ from attackers. Some terms for attackers are zeroed out if they have a weight smaller than $\varepsilon/n$. We can observe from Equation \ref{eq:bound_attacker} that the weight $w_j$ for a attacker is less than $\varepsilon/n$ if $\varepsilon'<2-\frac{1}{c} \ln{(\frac{1/\varepsilon-m/n}{1-m/n})} $. Hence, the error term due to some attackers can be dropped. The fourth line is obtained by dropping the error term due to the attacker and plugging in Equation \ref{eq:bound_benign}. The fifth line is simplification. The sixth line drops the $m$ term in the denominator of the previous line. Note that the effect of dropping the $m$ term would be minimal if we have a large $c$. 

Here we have finished the proof on the error bound of the robust estimate. The bound can be tighter if we do not drop the $m$ term in the last line. However, we present the current looser bound for readability. Next, we are going to prove the condition for improving the robust estimate.

In the best case where we have an arbitrarily small $\varepsilon'$, our algorithm can at best achieve an error bounded of
\begin{equation}\label{eq:best_case}
\begin{split}
        &\frac{n}{(n-m)e^{c(2-\varepsilon')} + m} \max_{l,w_l\geq \varepsilon/n}{\left(\|\vx_l\|\right)} \\&\mathrel{\stackrel{\makebox[0pt]{\mbox{\normalfont\tiny $\varepsilon'\to 0$}}}{\quad\longrightarrow\quad}} \frac{1}{(1-m/n)e^{2c}+m/n} \max_{l,w_l\geq \varepsilon/n}{\left(\|\vx_l\|\right)}.
\end{split}
\end{equation}

The higher the fraction of attackers, the larger the best error bound we can achieve. If there are less than 50\% attackers, the upper bound of the relative error $\frac{\|\vq_t-\vmu_{robust}\|}{\max_i{\|\vx_i\|}}$ is controlled by $2e^{-2c}$, which is about $10^{-9}$ for $c=10$.

In general, if we want our algorithm to give a better approximation than the previous iteration (with a rate $\gamma<1$), then we require
\begin{equation*}
    \begin{split}
        &\max{\left(e^{c\varepsilon'}-1,  \frac{n}{(n-m)e^{c(2-\varepsilon')} + m}\right)} \max_{l,w_l\geq \varepsilon/n}{\left(\|\vx_l\|\right)}\\ &\quad\leq \gamma \|\vq_{t-1}-\vmu_{robust}\|\\&\quad\leq \gamma\delta \max_l{\left(\|\vx_l\|\right)}. 
    \end{split}
\end{equation*}
 The second inequality is the condition on $q_{t-1}$ in Equation \ref{eq:condition}.

Hence, we require 
\begin{equation}\label{eq:convergence_condition}
\begin{split}
        e^{c\varepsilon'}-1 &\leq \gamma\delta\\
        \varepsilon' &\leq \frac{1}{c}\ln(\gamma\delta+1)= \frac{\gamma\delta}{c}+O(\gamma^2\delta^2)
\end{split}
\end{equation}
and
\begin{equation}\label{eq:convergence_condition2}
\begin{split}
         \frac{n}{(n-m)e^{c(2-\varepsilon')} + m} &\leq \gamma\delta\\
        \varepsilon' &\leq 2-\frac{1}{c}\ln{\left(\frac{\gamma^{-1}\delta^{-1}-m/n}{1-m/n}\right)}
\end{split}
\end{equation}

The equality in (\ref{eq:convergence_condition}) is due to Taylor expansion.

It means that we require our attention module to approximate the similarity measure $h'(k,q)$ with an error of at most $\frac{\gamma\delta}{c}$ if we have a large $c$. For example, we require $\varepsilon'\leq 1\%$ if we want to bound the relative error to $10\%$ in one iteration. 
\end{proof}

In practice, the true similarity measure $h_{*}(k,q)$ that identifies the attackers may not be the same as the $h'(k,q)$ we approximate based on the training data. In this case, the error term $\varepsilon'$ may not be small. The estimation could have a larger error according to the bound in the first part of Lemma \ref{lem1} and it may not improve throughout iterations since it may violate the condition in the second part of Lemma \ref{lem1}. Moreover, the higher the fraction of attackers, the worse we can do in the best case according to the bound. Fortunately, the first part of Lemma \ref{lem1} suggests that using a smaller scale factor $c$ may still give a good approximation even the similarity measure was not approximated well. In the extreme case when $c=0$, our approximation reduces to simple mean. In other words, when the attackers behave differently from what we simulated in the training (such as a new attack), using a conservative value of the scale factor $c$ could prevent false detection of attacker and false rejection of benign client.